%% file: main.tex
\documentclass{llncs}

\usepackage{eccv}

\usepackage{eccvabbrv}

\usepackage{multirow}

\usepackage{graphicx}
\usepackage{booktabs}

\usepackage[accsupp]{axessibility}  

\usepackage{hyperref}

\usepackage{orcidlink}

\usepackage[table]{xcolor}
\usepackage{float}
\usepackage{wrapfig}
\usepackage{graphbox}
\usepackage{siunitx}
\usepackage{array}
\usepackage{listings}
\usepackage{subcaption}
\usepackage{makecell}
\newcommand{\eqcontrib}{$^\ast$}
\newcommand{\corresponding}{$^\dagger$}

\begin{document}
\title{R3DP: Real-Time 3D-Aware Policy for Embodied Manipulation}

\author{\small 
Yuhao Zhang\inst{1,2}\eqcontrib
\and Wanxi Dong\inst{3,4}\eqcontrib
\and Yue Shi\inst{1}\eqcontrib
\and Yi Liang\inst{1}\eqcontrib
\and Jingnan Gao\inst{1} \and
\newline Qiaochu Yang\inst{1,4}
\and Yaxing Lyu\inst{7}
\and Zhixuan Liang\inst{5}
\and Yibin Liu\inst{8}
\and Congsheng Xu\inst{1}
\newline Xianda Guo\inst{6,2}
\and Wei Sui\inst{2}
\and Yaohui Jin\inst{1}
\and Xiaokang Yang\inst{1}
\and Yanyan Xu\inst{1}
\and Yao Mu\inst{1}\corresponding
}

\authorrunning{ }
\institute{
Shanghai Jiao Tong University
\and D-Robotics
\and Southern University of Science and Technology
\and Xspark AI
\and The University of Hong Kong
\and Wuhan University
\newline \and Xiamen University Malaysia
\and Northeastern University
\newline \email{\{yuhao\_zh, muyao\}@sjtu.edu.cn}
}

\pagestyle{plain}

\maketitle
\begin{center}
\vspace{-0.4cm}
\url{https://dazazh.github.io/r3dp-project-page/} 
\end{center}
\renewcommand{\thefootnote}{}
\footnotetext{$^\ast$ Co-first authors, equal contribution. $^\dagger$ Corresponding author. }

\renewcommand{\thefootnote}{\arabic{footnote}}  

\input{R3DP/sec/0_abstract}
\input{R3DP/sec/1_intro}
\input{R3DP/sec/2_related}

\input{R3DP/sec/3_preliminary}
\input{R3DP/sec/4_method}
\input{R3DP/sec/5_exp}
\input{R3DP/sec/6_conclusion}

\bibliographystyle{splncs04}
\bibliography{main}

\input{R3DP/sec/X_suppl}

\end{document}

%% file: R3DP/sec/0_abstract.tex
\begin{abstract}

\vspace{-15pt}
Embodied manipulation requires accurate 3D understanding of objects and their spatial relations to plan and execute contact-rich actions. 
While large-scale 3D foundation models provide strong priors, their computational cost incurs prohibitive latency for real-time control. 
We propose \textbf{R}eal-time \textbf{3D}-aware \textbf{P}olicy (\textbf{R3DP}), which integrates powerful 3D priors into manipulation policies without sacrificing real-time performance.
A core innovation of R3DP is the asynchronous fast–slow collaboration module, which sophisticatedly integrates large-scale 3D model's priors into the policy without compromising real-time performance. The system queries the pre-trained slow system (VGGT) only on sparse key frames, while simultaneously employing a lightweight Temporal Feature Prediction Network (TFPNet) to predict features for all intermediate frames. By leveraging historical data to exploit temporal correlations, TFPNet explicitly improves task success rates through consistent feature estimation. Additionally, we introduce a Multi-View Feature Fuser (MVFF) that aggregates features across views by explicitly incorporating camera intrinsics and extrinsics. R3DP offers a plug-and-play solution for integrating 3D foundation models into real-time inference systems. 
We evaluate R3DP against multiple baselines across different visual configurations. R3DP effectively harnesses large-scale 3D priors to achieve superior results, outperforming single-view and multi-view DP by 32.9\% and 51.4\% in average success rates. Furthermore, by decoupling heavy 3D understanding from policy execution, R3DP achieves a 44.8\% reduction in inference time compared to a naive DP+VGGT integration.

\keywords{Imitation Learning \and Diffusion Policy \and 3D Foundation Model}
\end{abstract}

%% file: R3DP/sec/1_intro.tex
\section{Introduction}


\label{sec:intro}
Humans naturally perceive and reason about the world in 3D or 4D, leveraging spatial perception and temporal continuity to perform complex manipulation tasks. 
However, current learning-based imitation methods \cite{chi2025diffusion, shukor2025smolvla, pearce2023imitating} for embodied manipulation often \textit{i}) lack an explicit, native 3D representation of the scene, relying primarily on 2D visual features that are weakly grounded in geometry, and \textit{ii}) struggle to effectively and efficiently model temporal context from historical observations, achieving limited success.

A straightforward way to compensate for the lack of 3D awareness is to augment RGB observations with additional modalities such as depth maps or point clouds. Prior works~\cite {manuelli2019kpam, zeng2020transporter} along these lines have shown that the injection of 3D information can indeed improve manipulation performance. However, the reliance on extra sensors or calibrated rigs of these methods increases system complexity. Such dependency leaves the system vulnerable to noise on transparent, reflective, or textureless surfaces, thereby limiting their practicality for real-world large-scale deployment. Moreover, embodied manipulation is inherently sequential that interaction trajectories naturally contain rich temporal cues about contacts, occlusions, and object motion. Several approaches attempt to exploit this by adding recurrent modules or memory mechanisms~\cite{nair2022r3m, o2024open} to imitation policies. But temporal context is often treated as an implicit hidden state, making it difficult to enforce spatial–temporal consistency or maintain a stable 3D understanding of the scene over long horizons.  
\begin{figure}[t]
    \centering
    \includegraphics[width=0.9\linewidth]{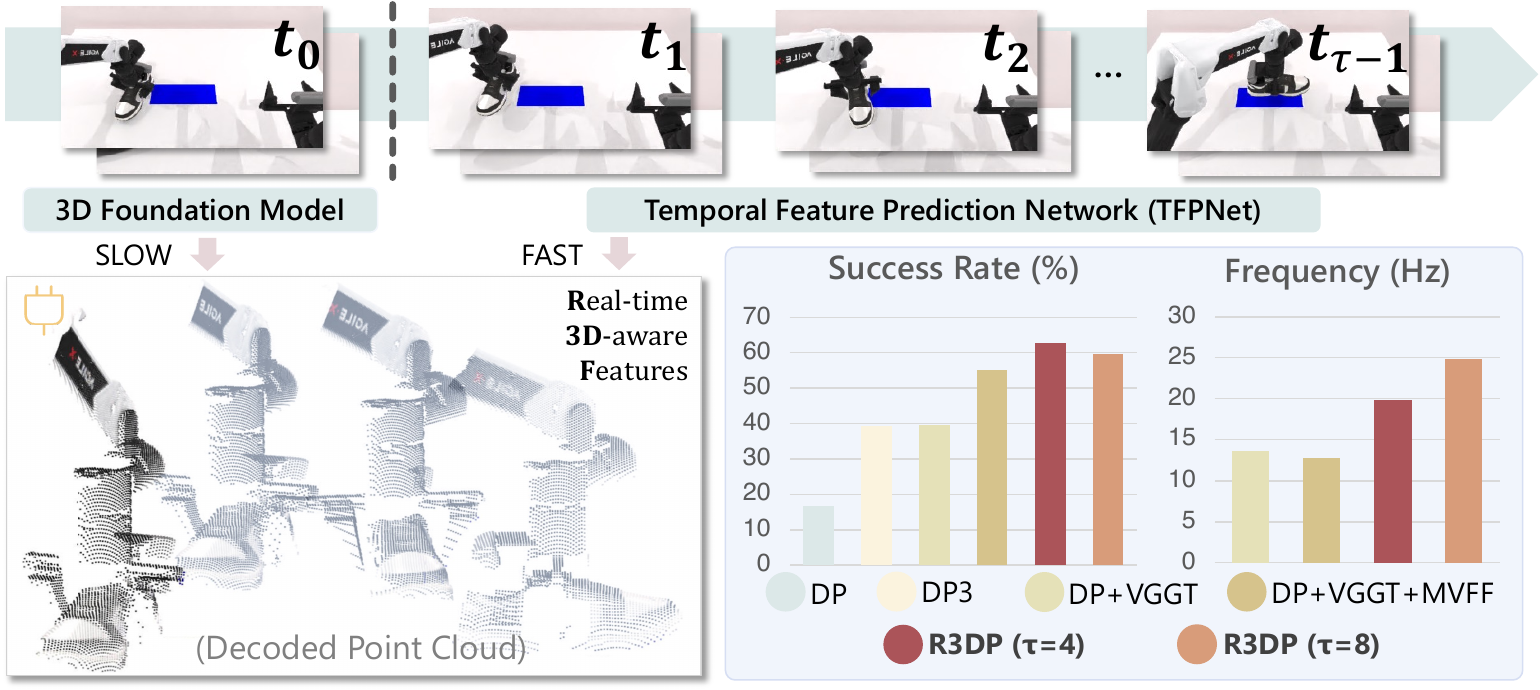}
    \vspace{-5pt}
    \caption{Key modules and performance of R3DP. Our framework explicitly integrates 3D priors from large-scale foundation models (e.g., VGGT) via an asynchronous fast-slow collaboration mechanism. Overall, R3DP achieves real-time, 3D-aware inference, significantly improving both manipulation success rates and processing frequency.}
    \vspace{-0.5cm}
    \label{fig:teaser_wrap}
\end{figure}

Recent progress has seen large-scale 3D vision foundation models \cite{vggt, depthany} rapidly develop, which presents good opportunities to obtain 3D priors directly from RGB inputs. These models offer a promising path to endow embodied policies with 3D spatial understanding, and without requiring additional depth or point cloud sensors. Unfortunately, their high computational cost makes per-frame inference far too slow for real-time robot control, especially in multi-view settings. Therefore, it is essential to design an approach that \textit{effectively and efficiently integrates these large-scale 3D priors with temporal context} into embodied manipulation policies, enabling 3D-aware control under real-time constraints.

We propose R3DP, a real-time 3D-aware policy for embodied manipulation that aggregates both 3D priors and temporal context information. An overview of R3DP and its performance are shown in Fig.~\ref{fig:teaser_wrap}. Basically, we integrate the 3D intermediate features from the large 3D foundation model VGGT~\cite{vggt} into the diffusion policy~\cite{chi2025diffusion}. However, running such a large 3D model on every frame is prohibitively slow. Therefore, to satisfy real-time constraints, we introduce an Asynchronous Fast–Slow Collaboration (AFSC) module with a lightweight Temporal Feature Prediction Network (TFPNet) for fast feature inference. AFSC queries VGGT only on sparse key frames and quickly derives features for the other frames using TFPNet. TFPNet is pre-trained to predict the current frame from historical frames, not only helping AFSC satisfy the requirement of real-time 3D prior injection but also providing additional temporal context to the policy. Furthermore, in addition to leveraging single-view 3D priors, we introduce a dedicated Multi-View Feature Fuser (MVFF) to explicitly aggregate multi-view features into a consistent 3D representation. MVFF leverages camera intrinsics and extrinsics to guide cross-view fusion, inspired by~\cite{lin2025sem, li2025cameras}.

The contributions of this work can be summarized as follows:
\begin{itemize}

\item \textbf{A Plug-and-Play 3D-Aware Policy Paradigm:} We propose R3DP, a real-time framework that seamlessly integrates 3D foundation model features into 2D image-based policies. This architecture endows robots with robust spatial perception while preserving the inherent strengths of 2D policy.

\item \textbf{Efficient Asynchronous Inference Mechanism:} To bypass 3D computational bottlenecks, we introduce an Asynchronous Fast–Slow Collaboration (AFSC) system, sophisticatedly integrating high-latency 3D features into real-time policies. This fast-slow collaboration ensures robust responsiveness during dynamic manipulation tasks.

\item \textbf{Geometry-Informed Multi-View Feature Fusion:} We introduce a Multi-View Feature Fuser (MVFF) that explicitly leverages camera intrinsics and extrinsics to aggregate features across multiple perspectives. This ensures spatial consistency and enhances the policy’s robustness in complex environments with occlusions.

\item \textbf{Extensive Validation of Effectiveness and Efficiency:} We extensively evaluate R3DP across diverse tasks, visual configurations, and baselines. R3DP improves average success by 32.9\% and 51.4\% over single-view and multi-view DP, respectively, and reduces inference latency by 44.8\% compared to the naive DP+VGGT integration via asynchronous fast–slow pipeline.


\end{itemize}

%% file: R3DP/sec/2_related.tex
\section{Related Work}

\noindent\textbf{2D-based Imitation Learning.}
Recent advances in imitation learning have demonstrated strong policy performance using purely 2D visual inputs.
Notable approaches~\cite{zhao2023learning, chi2025diffusion, black2410pi0, black2025pi_, liu2024rdt, gong2025carp, ma2025cdp} leverage powerful visual encoders and generative policies to learn from demonstrations. While effective on many manipulation tasks, these methods typically rely on RGB observations rather than explicitly reasoning about the underlying 3D scene. As a result, they often struggle in tasks requiring spatial reasoning, depth understanding, or handling occlusions. Moreover, multi-view inputs are often fused via simple concatenation or flattening, without leveraging camera geometry or enforcing spatial consistency.

\noindent\textbf{3D-Aware Imitation Learning.}
To overcome the limitations of 2D-only perception, several methods~\cite{Ze2024DP3, ze2024generalizable, lu2025h, ke20243d, lin2025sem, shridhar2023perceiver, grotz2024peract2} use RGB-D, point cloud representations, or voxel-based representations to enhance spatial awareness in robotic tasks. However, such approaches require additional depth sensors, which are prone to noise, particularly in challenging conditions, such as those involving transparent or reflective surfaces. Furthermore, real-world deployment may be constrained by hardware costs or limited sensor coverage. In contrast, our work avoids explicit depth sensing and instead builds upon 3D visual foundation models to extract geometry-aware features directly from RGB inputs.

\noindent\textbf{Imitation Policies with Large-scale 3D Prior.}
Recent methods have explored using pre-trained 3D foundation models~\cite{fast3r,vggt,dens3r,cut3r,flare,monst3r,mvdust3r,must3r,pow3r,pi3,dust3r,mast3r,wang2024spann3r,MoRE} to provide strong priors for manipulation. These methods~\cite{lin2025evo, abouzeid2025geoaware, qian2025gp3} simply fused pre-trained features into imitation policies, whereas methods like eVGGT~\cite{dinh2025improving} use distillation to make VGGT efficient for robotic policy inference. These approaches do enhance spatial understanding and geometry-aware action generation. However, integrating such large-scale priors introduces significant challenges: these models often suffer from \textit{scale shifting}—a mismatch in feature magnitudes over time that degrades policy stability. More critically, these models typically incur significant \textit{inference latency} in real-world deployment. Furthermore, multi-view information is often merged heuristically, without explicit encoding of camera intrinsics or extrinsics. To address these issues while maintaining spatial-temporal consistency, we introduce a \textit{fast-slow system} inspired by prior temporal models~\cite{feichtenhofer2019slowfast}, which explicitly separates transient local features from relatively stable 3D scene representations, enabling the policy to maintain geometric consistency while preserving real-time performance.

%% file: R3DP/sec/3_preliminary.tex
\section{Preliminary}
\noindent\textbf{Imitation Learning.} Imitation learning aims to acquire a control policy by mimicking expert behavior from demonstrations. 
Given an expert dataset $\mathcal{D} = \{(s_i, a_i, s'_i)\}_{i=1}^N$, where $s_i \in S$ is the state, $a_i \in A$ is the action taken in state $s_i$, and $s'_i$ is the resulting next state, the goal of imitation learning is to obtain a policy $\pi: S \rightarrow A$ that reproduces the expert’s behavior. 

\noindent\textbf{Diffusion Policy.} Diffusion policy~\cite{chi2025diffusion} (DP) is a visuomotor imitation learning method that models the distribution of future action sequences with a conditional Denoising Diffusion Probabilistic Model (DDPM). Instead of directly regressing actions in an autoregressive manner, DP learns to iteratively denoise a noisy action sequence into a feasible plan conditioned on the current observation. Concretely, at time step $t$, the policy conditions on an observation $O_t$ (e.g., stacked RGB frames and proprioception) and predicts a future action sequence $A_t^0$. DP trains a parameterized noise prediction network \(\epsilon_\theta\) and applies it in a $k$-step iterative denoising process:
\begin{equation}
A_t^{k-1} = \alpha(A_t^k - \gamma \epsilon_\theta(O_t, A_t^k, k) + \mathcal{N}(0, \sigma^2 I)),
\end{equation}
where \(A_t^k\) is the randomly sampled Gaussian noise, \(A_t^0\) is the expected noise-free action output at time \(t\), \(O_t\) is the observation data used as the condition, \(\gamma\) is the learning rate, and \(N(0, \sigma^2 I)\) represents the Gaussian noise added during the iteration. To obtain the optimal parameters \(\theta\), the training loss is defined as:
\begin{equation}
    \mathcal{L}_{DP}
    =
    \mathbb{E}_{(O_t, A_t^{0}),\, k,\, \boldsymbol{\epsilon}}
    \bigl[
        \lVert
            \boldsymbol{\epsilon}
            -
            \epsilon_\theta(O_t, A_t^{0} + \boldsymbol{\epsilon}_k, k)
        \rVert_2^2
    \bigr],
\end{equation}
where \(k\) is the randomly chosen denoising iteration step, and \(\epsilon^k\) is the random noise with a variance matching the \(k\)-step sampling.

\noindent\textbf{Projective Positional Encoding (PRoPE).}
Existing methods encode only SE(3) extrinsic pose $T^{cw}_i$ , fundamentally neglecting intrinsic parameters $K_i$ and failing on cameras with varying focal lengths. Projective Positional Encoding (PRoPE)~\cite{li2025cameras} addresses this by modeling complete frustum geometry via relative projective transformations.
The core innovation of PRoPE lies in modeling inter-camera relationships through relative projective transformations:
{\scriptsize\begin{equation}
\tilde{\mathbf{P}}_{i,j}=\begin{bmatrix}\mathbf{K}_i&\mathbf{0}\\\mathbf{0}&1\end{bmatrix}\textbf{T}^{cw}_i(\textbf{T}^{cw}_j)^{-1}\begin{bmatrix}\mathbf{K}_j^{-1}&\mathbf{0}\\\mathbf{0}&1\end{bmatrix}.
\end{equation}}
PRoPE is injected via GTA-style~\cite{miyato2023gta} attention using per-token block-diagonal transformation:
{\scriptsize
\begin{gather}
    D_t^{\text{PRoPE}} = 
\begin{bmatrix}
D_t^{\text{Proj}} & \mathbf{0} \\
\mathbf{0} & D_t^{\text{RoPE}}
\end{bmatrix} \in \mathbb{R}^{d \times d}, \\
D_t^{\text{Proj}} = I_{d/8} \otimes \tilde{P}_{i(t)} \in \mathbb{R}^{\frac{d}{2} \times \frac{d}{2}},\\
D_t^{\text{RoPE}} = 
\begin{bmatrix}
\text{RoPE}_{d/4}(x_t) & \mathbf{0} \\
\mathbf{0} & \text{RoPE}_{d/4}(y_t)
\end{bmatrix} \in \mathbb{R}^{\frac{d}{2} \times \frac{d}{2}}.
\end{gather}}

This formulation guarantees global frame invariance, reduces to SE(3) encoding when intrinsics are shared, and degrades gracefully to RoPE for tokens within identical views.

%% file: R3DP/sec/4_method.tex
\begin{figure}[t]
    \centering
    \includegraphics[width=\linewidth]{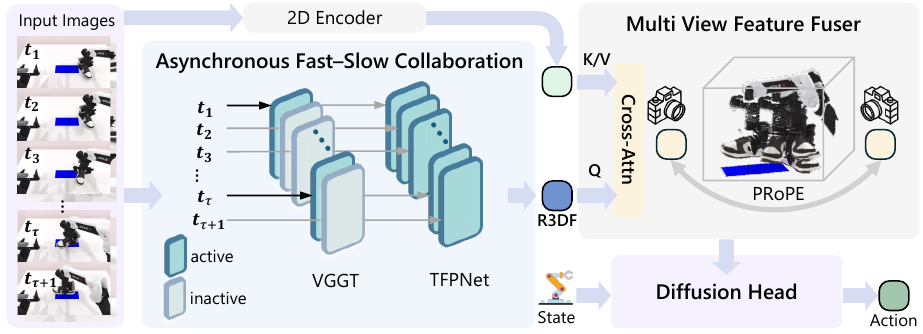}
    \caption{Overview of the R3DP architecture. 
    R3DP serves as a 3D-aware perception module that seamlessly replaces visual encoders in existing imitation learning frameworks. Within the AFSC module (\ref{41}), sparse keyframes are processed by a 3D foundation model (VGGT), while intermediate frames are handled by our TFPNet (\ref{sec:tfp}) for real-time temporal reasoning. 
    MVFF module (\ref{sec:mvff}) leverages cross-attention with PRoPE to fuse 2D-3D features into consistent multi-view representations for control.}
    \vspace{-0.5cm}
    \label{fig:pipeline}
\end{figure}
\section{Methodology}
Although diffusion policy is powerful in modeling complex action distributions, it faces several key limitations when directly applied to 3D robotic manipulation: \textbf{\textit{i}) Lack of 3D perception.} Standard diffusion policy and its variants typically rely on generic 2D visual encoders (e.g., ResNet) to extract observation features, and thus lack explicit 3D spatial perception of the scene; \textbf{\textit{ii}) Implicit multi-view fusion.} Standard implementation of diffusion policy uses simple feature concatenation for implicit fusion of multiple views. This implicit fusion does not explicitly encode camera intrinsics and extrinsics, and therefore underutilizes the underlying geometric structure across views.

The overall architecture of R3DP is illustrated in Figure~\ref{fig:pipeline}. Compared to standard 2D image-based imitation policies, R3DP is designed around several key aspects: 3D-aware perception, real-time feasibility, temporal context assistance, and explicit multi-view fusion. 
Concretely, it consists of three core components. First, the Asynchronous Fast-Slow Collaboration (AFSC) module injects 3D priors into the policy by leveraging a large 3D vision foundation model (VGGT) as a slow branch that processes sparse key frames and produces 3D-aware features. 
Second, the lightweight Temporal Feature Prediction Network (TFPNet) acts as the fast branch. It is pre-trained to predict current-frame features from historical observations, both propagating VGGT features to intermediate frames and providing explicit temporal modeling in real time. 
Third, a Multi-View Feature Fuser (MVFF) explicitly aggregates per-view features using camera intrinsics and extrinsics to form a consistent 3D-aware representation across viewpoints.

\subsection{Asynchronous Fast-Slow Collaboration}
\label{41}
Directly applying large 3D vision models at every frame to extract 3D features cannot meet the real-time control requirements of embodied manipulation. To enable efficient 3D perception, we propose an AFSC module that collaborates a slow 3D backbone with a fast temporal predictor to predict 3D features.


AFSC treats a large 3D vision foundation model as the slow system that takes current multi-view RGB images every $\tau$ inferences, and computes high-fidelity Slow 3D-aware Features (S3DF). Meanwhile, AFSC takes a lightweight Temporal Feature Prediction Network (TFPNet) as the fast system that propagates these features to every intermediate frame using historical context to predict the Real-time 3D-aware Features (R3DF).
Specifically, the slow and fast systems $\Phi_{\text{slow}}$ and $\Phi_{\text{fast}}$, can be represented as:

{\small
\begin{equation}
\begin{aligned}
\Phi_{\text{slow}}:\ \mathbf{I}_{i_0}\ &\overset{\text{VGGT}}{\longmapsto}\ F^\mathrm{S3D}_{i_0}, \\
\Phi_{\text{fast}}:\ (F^\mathrm{R3D}_{i-1}\ ,\mathbf{I}_i)\ &\overset{\text{TFPNet}}{\longmapsto}\ F^\mathrm{R3D}_i,
\end{aligned}
\end{equation}}
where $i\in(i_0, i_0+\tau)$ is the current inference iteration, $\mathbf{I}_{i}\in\mathbb{R}^{V\times 3\times H\times W}$ is multi-view RGB frame of iteration $i$, $F^\mathrm{S3D}_{i}$, $F^\mathrm{R3D}_{i}$ are slow and real-time features, respectively. In this way, AFSC efficiently injects 3D priors to meet the real-time requirements of embodied manipulation.

After obtaining per-view feature, we explicitly aggregate features across views (as in Sec. \ref{sec:mvff}), yielding a fused 3D-aware visual representation. We feed the derived features into the diffusion denoiser to generate the future action sequence. To expose longer temporal dependencies during training while keeping computation affordable, we modify the diffusion policy sampler so that each training tuple contains eight frames uniformly spaced by a fixed stride of eight timesteps (instead of three consecutive frames). This yields an effective temporal window of $8 \times 8 = 64$ frames, suitable for common manipulation horizons and aligned with AFSC's single-key-frame schedule. Within each tuple, the first frame is processed by VGGT to produce the initial S3DF; subsequent frames are updated by TFPNet to produce ordered R3DF. Training is performed serially within each tuple but in parallel across batches. In this stage, gradients are only updated for the diffusion head, as the vision backbones (VGGT and TFPNet) remain frozen, and the entire framework is trained end-to-end with a unified diffusion loss.





\subsection{Temporal Feature Prediction Network}
\label{sec:tfp}
\begin{wrapfigure}[18]{r}{0.5\textwidth} 
    \vspace{-0.8cm}
    \centering
    \includegraphics[width=\linewidth]{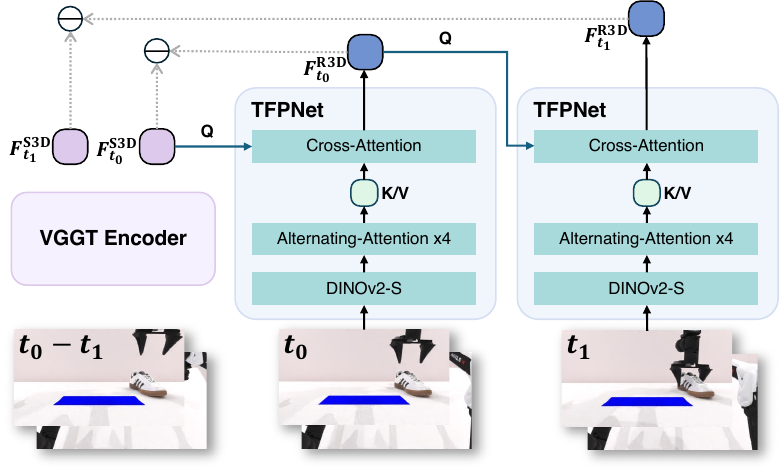} 
    \caption{Architecture and training objective of TFPNet. For clarity, we show the unrolled structure for the first two timesteps; in practice, the network is trained over a sequence of four timesteps. TFPNet leverages historical information to augment current observations, enabling 3D-aware control with real-time inference efficiency.}
    \label{fig:tfpnet}
\end{wrapfigure}

To bridge the gap between 3D vision backbones and real-time embodied manipulation, we pre-train a lightweight Temporal Feature Prediction Network (TFPNet) that leverages historical-frame 3D-aware features to predict the current-frame feature online in real time. 

TFPNet is trained by distilling supervision from a large 3D vision foundation model (VGGT). For each episode, we sample 4 frames $\{t_0,t_1,t_2,t_3\}$ in chronological order, with a random interval of 1--8 frames between them to simulate the temporal gap that typically occurs between two inference steps in manipulation (i.e., after an action chunk is executed). 
These ordered frames form one training tuple. During training, all frames are first processed by VGGT to obtain the ground-truth 3D features $F^\mathrm{S3D}$, which are the unified tokens extracted from 24 layers for VGGT's dense output as follows:
\begin{equation}
    F^\mathrm{S3D} = \text{VGGT}(\mathbf{I}_{t_0}).
\end{equation}

The TFPNet then receives the VGGT feature from the first frame and iteratively predicts features for subsequent frames. Then R3DF is predicted by the TFPNet $\mathcal{G}$ of frame $t$ as follows:
{\small
\begin{equation}
F^\mathrm{R3D}_{t_k} =
\begin{cases}
\mathcal{G}(F^\mathrm{S3D}, \mathbf{I}_{t_0}), & k=0,\\[6pt]
\mathcal{G}(F^\mathrm{R3D}_{t_{k-1}},  \mathbf{I}_{t_k}), & k\in\{1,2,3\},
\end{cases}
\label{eq:fastsystem}
\end{equation}}

\noindent where $\mathbf{I}_{t}\in\mathbb{R}^{V\times 3\times H\times W}$ is multi-view RGB view of timestep $t$, $F^\mathrm{S3D}_{t}$, $F^\mathrm{R3D}_{t}$ are slow and real-time features of timestep $t$. 

As shown in Figure~\ref{fig:tfpnet}, TFPNet follows a compact, 3D-aware design. 
Each view is first encoded by a DINOv2-S\cite{oquab2023dinov2} backbone to produce 2D features, and then passed through four Alternating-Attention (AA) Transformer~\cite{vggt} blocks to perform cross-view aggregation.
Finally, the aggregated per-view tokens attend to the historical feature (e.g., $F^\mathrm{R3D}_{t-1}$) via cross-attention, injecting temporal priors to stabilize optimization and improve long-horizon reasoning.


The TFPNet is pre-trained using a cosine similarity loss between the predicted features $f_t$ and the VGGT features $F_t$ as follows:
{\small
\begin{equation}
\mathcal{L}_\mathrm{TFP} = \sum_{i=0}^{3} \big( 1 - \cos(F^\mathrm{S3D}_t, F^\mathrm{R3D}_t) \big).
\label{eq:loss}
\end{equation}}
This objective encourages TFPNet to produce more accurate features leveraging temporal context. Besides, the lightweight design of TFPNet enables real-time inference. After pre-training, TFPNet is inserted into our AFSC module as the fast system, enabling real-time inference.

\subsection{Multi-View Feature Fuser}
\label{sec:mvff}
The cameras mounted on the robot wrist or fixed in the background can inherently provide ground-truth extrinsic parameters. Inspired by this observation, we design the MVFF module to leverage these parameters for enhancing multi-view 3D representations. 
Although the VGGT backbone can implicitly infer camera parameters from its feature representations, the estimated values remain imperfect. To fully exploit the reliable ground-truth camera information, the MVFF module explicitly injects camera parameters into the model, thereby achieving a more consistent and effective fusion of multi-view 2D and 3D features.

The VGGT feature serves as a unified multi-view representation; however, it still preserves view-specific properties, since each view can be independently decoded into a distinct depth map. In parallel, each view’s image is processed by a ResNet encoder to obtain 2D features.
In detail, We first apply cross-attention to fuse the 2D and 3D representations for each view, obtaining a per-view 2D--3D hybrid feature. 
Then, a PRoPE module, which follows the RoPE-style relative positional encoding to avoid dependence on absolute order, explicitly incorporates camera intrinsics and extrinsics to enhance multi-view feature aggregation. Formally, the overall process is expressed as:
{\small
\begin{gather}
Q=W_QF_t^{2\text{D}}, K=W_KF^\mathrm{R3D}_{t}, V=W_VF^\mathrm{R3D}_{t},\\
\tilde{F}_t = \text{softmax}(\frac{QK^\top}{\sqrt{d}})V,\\
F_t^{\text{fused}} = \mathrm{PRoPE}\!\left(\tilde{F}_t, K_t, E_t\right),
\end{gather}}
where $K_t$ and $E_t$ denote the camera intrinsics and extrinsics, respectively.
The fused feature $F_t^{\text{fused}}$ serves as the final visual embedding that conditions the diffusion model during policy learning.

%% file: R3DP/sec/5_exp.tex
\section{Experiments}

\subsection{Experimental Setup on Simulation Benchmark}

\noindent\textbf{Simulated Environments.}  
We conduct experiments on 10 representative tasks of RoboTwin~\cite{chen2025robotwin}, a scalable dual-arm manipulation benchmark with diverse tasks and unified protocols. 
Among them, 9 tasks are drawn directly from the original RoboTwin benchmark. Meanwhile, we also include an additional task, \textit{Tube Insert}, to specifically evaluate the capability of 3D foundation model-based policies for handling transparent-object manipulation. Detailed task descriptions are provided in the supplementary materials.

\noindent\textbf{Baselines.}  
Our primary baseline is the Diffusion Policy (DP), evaluated under both single-view and multi-view settings. In the RoboTwin environment, the single-view configuration uses only the head camera, whereas the multi-view configuration uses both the head and front cameras.
We choose DP as the baseline because it is a lightweight visuomotor imitation model that relies solely on 2D image inputs and lacks explicit 3D perception or temporal modeling. This makes it an ideal testbed for assessing the benefits introduced by our R3DP framework. 

In addition, we include 3D Diffusion Policy (DP3)\cite{Ze2024DP3} as a depth-aware baseline. DP3 operates on head-view point cloud inputs and thus requires depth information and object bounding boxes for cropping. In simulation, where ground-truth depth and bounding boxes are available, DP3 performs competitively. However, its performance degrades significantly when depth is noisy or incomplete. To further evaluate robustness under realistic sensing conditions, we also compare Depth Anything v2 (DA2)\cite{yang2024depth} + DP3, where DA2 predicts depth maps that are back-projected into point clouds for DP3. 

We also introduce $\pi_0$~\cite{black2410pi0} as a representative VLA (Vision-Language-Action) baseline. Utilizing a large-scale VLM (Vision-Language Model) as its backbone and having been pre-trained on massive datasets, $\pi_0$ exhibits exceptional generalization capabilities. To align with the experimental settings of imitation policies, we fine-tuned a separate model for each task from the $\pi_0$-base using LoRA. 

\noindent\textbf{Evaluation Metrics.}  
To ensure a fair comparison across various models and tasks, all methods are evaluated using checkpoints from a consistent training stage under identical environmental configurations. For each task and each model, we evaluate the selected checkpoint in the RoboTwin environment and measure performance by computing the success rate over 100 execution trials. 

In addition, we measure the inference efficiency of each policy using the observation encoding time, defined as the average latency from raw multi-view images to the final fused feature used to condition the policy head. This metric captures the overall perception latency and serves as a key indicator for validating the inference speed advantage of our R3DP framework.

\noindent\textbf{Implementation Details.}  
For R3DP, we train the model with $\tau=8$ for 600 epochs on 4 NVIDIA 4090s using Adam (lr = 2e-5, batch size = 8). The universal TFPNet is trained on 8 NVIDIA A800s. For any part that involves the output of VGGT, we use bfloat16 precision to accelerate training and evaluation.
\begin{table}[t]
\centering
\footnotesize
\setlength{\tabcolsep}{3pt} 
\caption{Task success rates on RoboTwin benchmark. We report single-view Diffusion Policy (DP-single), multi-view Diffusion Policy (DP-multi), DP3, DP3+DA2, $\pi_0$ and Diffusion Policy with R3DP-enhanced feature (\textbf{R3DP}). For R3DP, we select $\tau=4$ and $\tau=8$. Note that $\tau$ can serve as a configurable hyperparameter at inference time to achieve a trade-off between inference speed and success rate.}
\resizebox{\textwidth}{!}{
\begin{tabular}{l ccc cccc}
\toprule
\textbf{Task} & DP-single & DP-multi & DP3 & DP3-DA2 &$\pi_0$ & \textbf{R3DP ($\tau$=4)} & \textbf{R3DP ($\tau$=8)} \\
\midrule
Block Hammer Beat    & 0\%    & 0\%    & 49\% & 7\% & 47\% & \cellcolor{gray!20}\textbf{77}\%   & \cellcolor{gray!20}\textbf{77}\% \\
Block Handover       & 1\%    & 2\%    & 48\% & 31\% & 71\% & \cellcolor{gray!20}\textbf{95}\%   & \cellcolor{gray!20}93\% \\
Bottle Adjust        & 55\%   & 6\%    & \textbf{75}\% & 37\% & 46\% & \cellcolor{gray!20}62\%   & \cellcolor{gray!20}56\% \\
Blocks Stack Easy    & 6\%    & 7\%    & 26\% & 3\% & \textbf{79}\% & \cellcolor{gray!20}69\%   & \cellcolor{gray!20}62\% \\
Shoe Place           & 45\%   & 19\%   & 49\% & 34\% & 71\% & \cellcolor{gray!20}\textbf{72}\%   & \cellcolor{gray!20}68\% \\
Container Place      & 48\%   & 14\%   & \textbf{89}\% & 37\% & 77\% & \cellcolor{gray!20}63\%   & \cellcolor{gray!20}54\% \\
Dual Shoes Place     & 6\%    & 9\%    & 11\% & 0\% & \textbf{29}\% & \cellcolor{gray!20}24\%   & \cellcolor{gray!20}20\% \\
Diverse Bottles Pick & 16\%   & 17\%   & 34\% & 7\% & \textbf{47}\% & \cellcolor{gray!20}31\%            & \cellcolor{gray!20}32\% \\
Put Apple Cabinet    & 92\%   & 38\%   & 98\% & 94\% & 64\% & \cellcolor{gray!20}\textbf{100}\%  & \cellcolor{gray!20}98\% \\
Tube Insert          & 92\%   & 64\%   & \textbf{97}\% & 32\% & 68\% & \cellcolor{gray!20}\textbf{97}\%   & \cellcolor{gray!20}\textbf{97}\% \\
\midrule
\textbf{Average}     & 36.1\% & 17.6\% & 57.6\% & 28.2\% & 59.9\% & \cellcolor{gray!20}\textbf{69.0\%} & \cellcolor{gray!20}65.7\% \\
\bottomrule
\end{tabular}
}
\label{tab:robotwin_dp_variants_64clean}
\end{table}

\begin{table}[t]
\centering
\setlength{\tabcolsep}{2pt} 
\caption{Average latency from raw multi-view images to fused features, averaged over four tasks (Blocks Stack Easy, Block Hammer Beat, Shoe Place, Diverse Bottles Pick) experimented in the ablation study.}
\vspace{-0.2cm}
{\scriptsize
\begin{tabular}{l|cccc}
\toprule
\textbf{Latency (ms $\downarrow$)} & DP+VGGT & DP+VGGT+MVFF & \textbf{R3DP($\tau=4$)} & \textbf{R3DP($\tau=8$)} \\
\midrule
Obs. Encoder & 73.1 & 78.3 ({\color{red}$\uparrow7.1\%$}) & \cellcolor{gray!20}50.5 ({\color{green!90!black}$\downarrow30.9\%$}) & \cellcolor{gray!20}\textbf{40.3} ({\color{green!90!black}$\downarrow44.8\%$}) \\
Action Expert & 56.6 & 57.7 & \cellcolor{gray!20}55.3 & \cellcolor{gray!20}56.7 \\
\bottomrule
\end{tabular}}
\label{frequency_results_transposed}
\vspace{-0.5cm}
\end{table}
\subsection{Results on Simulation Benchmark}
\noindent\textbf{Quantitative Results.}
We first compared R3DP against two simple yet efficient policies: the single-view Diffusion Policy and the multi-view Diffusion Policy. As shown in Table~\ref{tab:robotwin_dp_variants_64clean}, R3DP achieves consistent and remarkable performance gains across all ten manipulation tasks in the RoboTwin benchmark. 
In tasks requiring fine-grained 3D reasoning and precise coordination—such as \textit{Block Hammer Beat}, \textit{Blocks Stack Easy}, and \textit{Block Handover}—R3DP attains success rates of 77\%, 69\%, and 95\%, respectively, representing dramatic improvements over the single-view Diffusion Policy baseline (0\%, 6\%, and 1\%). Even in simpler placement or alignment tasks, including \textit{Shoe Place } and \textit{Container Place}, our framework consistently improves accuracy by more than 15\%. 

Interestingly, we observe that the Diffusion Policy with multi-view inputs (denoted as DP-multi) exhibits lower success rates than the single-view DP. This indicates that traditional Diffusion Policy struggles to effectively fuse multi-view information, as its naïve combination of multiple perspectives fails to leverage the rich 3D cues fully. In contrast, R3DP employs the MVFF (Multi-View Feature Fuser) module, which explicitly models cross-view interactions and effectively extracts 3D features from multi-view inputs.
Overall, R3DP achieves an average success rate of 69.0\%, outperforming the single-view and multi-view Diffusion Policies by 32.9\% and 51.4\%, respectively. 

We observed that the DP3 baseline achieves a significantly high success rate, which we hypothesize is due to its reliance on ground-truth depth data in simulation. However, when noisy point clouds generated by depth estimation models are used, its performance drops drastically. Compared to the DP3 baselines, our model surpasses DP3 and DP3 + DA2 by 11.4\% and 40.8\%. Moreover, we found that our model outperforms $\pi_0$ (with an average success rate of 59.9\%) on the majority of tasks, potentially because the limited data volume poses a significant challenge for a general-purpose VLA model to effectively converge on or fit the task-specific distributions.

\begin{figure}[t]
    \centering
    \setlength{\tabcolsep}{0pt}
    \renewcommand{\arraystretch}{0.0}

    \scalebox{0.75}{%
    {\footnotesize
    \begin{tabular}{cccccccc}
        & {\fontsize{7.5pt}{8.5pt}\selectfont\textbf{RGB}}
        & {\fontsize{7.5pt}{8.5pt}\selectfont\textbf{VGGT}}
        & {\fontsize{7.5pt}{8.5pt}\selectfont\textbf{Ours}}
        &
        & {\fontsize{7.5pt}{8.5pt}\selectfont\textbf{RGB}}
        & {\fontsize{7.5pt}{8.5pt}\selectfont\textbf{VGGT}}
        & {\fontsize{7.5pt}{8.5pt}\selectfont\textbf{Ours}}\\[2.5pt]

        \raisebox{8pt}{\rotatebox{90}{\textbf{Head}} \hspace{1.5pt}} &
        \includegraphics[width=0.2\textwidth]{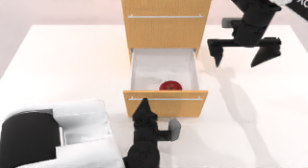} &
        \includegraphics[width=0.2\textwidth]{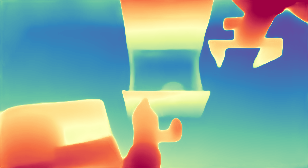} &
        \includegraphics[width=0.2\textwidth]{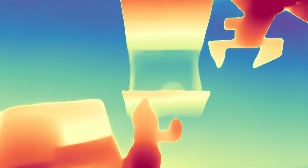} &
        \hspace{7pt}\raisebox{8pt}{\rotatebox{90}{\textbf{Head}} \hspace{1.5pt}} &
        \includegraphics[width=0.2\textwidth]{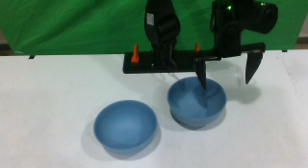} &
        \includegraphics[width=0.2\textwidth]{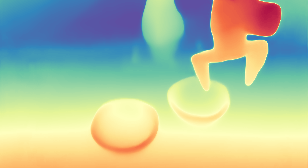} &
        \includegraphics[width=0.2\textwidth]{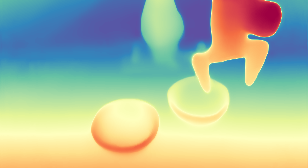} \\[0pt]
    
        \raisebox{8pt}{\rotatebox{90}{\textbf{Front}} \hspace{1.5pt}} &
        \includegraphics[width=0.2\textwidth]{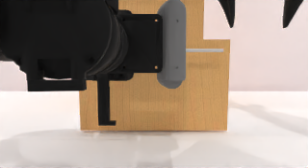} &
        \includegraphics[width=0.2\textwidth]{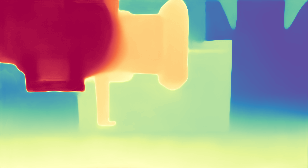} &
        \includegraphics[width=0.2\textwidth]{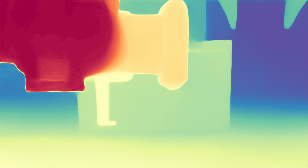} &
        \hspace{7pt}\raisebox{8pt}{\rotatebox{90}{\textbf{Front}} \hspace{1.5pt}} &
        \includegraphics[width=0.2\textwidth]{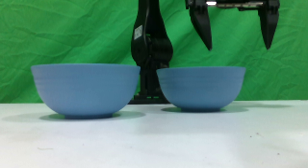} &
        \includegraphics[width=0.2\textwidth]{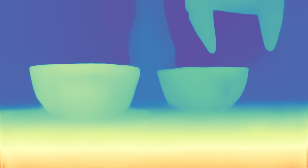} &
        \includegraphics[width=0.2\textwidth]{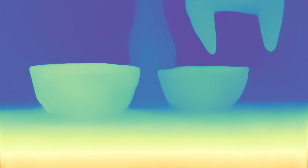} \\ [0.1cm]

        & \multicolumn{3}{c}{{\selectfont\textbf{Simulation}}}
        & \multicolumn{1}{c}{}
        & \multicolumn{3}{c}{{\selectfont\textbf{Real-World}}} \\[-0.3cm]
    \end{tabular}
    }
    }
    \caption{Visualization of depth maps decoded from VGGT features and from our TFPNet-predicted features passed through VGGT’s depth decoder. The close visual agreement indicates that our lightweight TFPNet effectively captures information generated by 3D foundation model in both simulation and real-world experiments.}
    \vspace{-0.8cm}
    \label{fig:tfpvis}
\end{figure}

These results validate the core novelty of R3DP: by asynchronously coupling a pretrained large-scale 3D backbone (VGGT) with a lightweight temporal prediction network, the policy can efficiently exploit rich geometric priors while still meeting real-time control constraints. Moreover, by visualizing the feature maps predicted by TFPNet and comparing them with those produced by VGGT in Figure~\ref{fig:tfpvis}, we observe highly consistent 3D representations, confirming that our lightweight network successfully captures VGGT-like 3D cues and thereby enables such performance gains with a much smaller model.

\noindent\textbf{Inference Speed.} 
We further evaluate the runtime efficiency of R3DP. 
While incorporating the fast--slow collaboration module inevitably introduces additional computation compared to the original DP, R3DP remains significantly faster than the naïve DP+VGGT combination, which queries the large 3D backbone at every frame. 
As shown in Table~\ref{frequency_results_transposed}, R3DP achieves real-time 3D feature processing latency at 40.3ms (vs. 73.1ms for DP+VGGT and 78.3ms for DP+VGGT+MVFF), 
demonstrating that our asynchronous design effectively balances accuracy and latency, enabling practical deployment in real-world robotic control.


\noindent\textbf{Configurable Efficiency and Robustness.}
R3DP supports real-time 3D feature processing with high flexibility. Our AFSC allows users to adjust the window size parameter $\tau$ to optimize inference latency based on hardware constraints. Notably, R3DP exhibits remarkable robustness to changes in $\tau$; for instance, when $\tau$ increases from 4 to 8, the encoding latency reduces by 20.2\% while maintaining a highly stable success rate (with a negligible fluctuation of only 3.3\%). This suggests that $\tau$ can be tuned at deployment to optimize real-time performance without retraining or sacrificing policy reliability. Full experiments on different $\tau$ are shown in the supplementary materials.

\subsection{Ablation Study}
\begin{table}[t]
  \centering
  \scriptsize
  \setlength{\tabcolsep}{7pt} 
  \caption{Ablation study on the effectiveness of key components. We progressively add the 3D foundation model (VGGT), Multi-View Feature Fusion (MVFF), and AFSC to the baseline (DP-multi). The final model with all components denotes \textbf{R3DP}.}
  \vspace{-0.2cm}
  \begin{tabular}{l|cccc} 
  \toprule
  \textbf{Task} & DP-multi & DP+VGGT & DP+VGGT+MVFF & \textbf{R3DP} \\ 
  \midrule
  Blocks Stack Easy    & 7\%  & 30\% & 51\% & \textbf{69\%} \\
  Block Hammer Beat    & 0\%  & 44\% & 74\% & \textbf{77\%} \\
  Shoe Place           & 19\% & 55\% & 66\% & \textbf{72\%} \\
  Diverse Bottles Pick & 17\% & 30\% & 30\% & \textbf{31\%} \\
  \midrule
  \textbf{Average} & 10.8\% & 39.8\% & 55.3\% & \textbf{62.3\%} \\
  \bottomrule
  \end{tabular}
  \vspace{-0.4cm} 
  \label{ablation:all_components}
\end{table}

The Table \ref{ablation:all_components} reports ablation results that validate the effectiveness of each key component in the proposed R3DP framework. 
We systematically examine the effectiveness of the 3D foundation model VGGT, the Multi-View Feature Fuser (MVFF), and the Asynchronous Fast–Slow Collaboration (AFSC) module.

\noindent\textbf{Effectiveness of 3D Foundation Model.}  
We first evaluate the impact of augmenting Diffusion Policy with the pretrained VGGT backbone, which we denote as DP+VGGT. 
In this setting, each frame is independently processed by VGGT without temporal modeling, allowing us to isolate the effect of 3D priors.  We use a  geometry-agnostic multi-level attention-based fusion module, rather than MVFF, to fuse 2D and 3D features.
Although no historical information or camera parameters are used, VGGT provides strong 3D-awareness that substantially improves spatial reasoning and manipulation precision.

As shown in Table~\ref{ablation:all_components}, using the 3D features obtained from VGGT (DP+VGGT) increased the average success rate of DP-multi from 10.8\% to 39.8\%. Specifically, for the geometry-aware tasks such as \textit{Blocks Stack Easy} and \textit{Block Hammer Beat}, the success rates of DP-multi were only 7\% and 0\%, respectively. After injecting VGGT features, the success rates of these two tasks increased to 30\% and 44\%, which is sufficient to demonstrate that 3D foundation models can serve as powerful sources of geometric priors to enhance policy perception.

\noindent\textbf{Effectiveness of MVFF.}  
Building on the VGGT-enhanced baseline, we further introduce the Multi-View Feature Fuser (MVFF) to explicitly leverage camera intrinsics and extrinsics for geometry-aware cross-view feature alignment, enabling the policy to more effectively integrate complementary visual cues across different viewpoints.
As shown in Table~\ref{ablation:all_components}, compared to DP+VGGT, integrating MVFF (DP+VGGT+MVFF) yields a clear performance gain, with the average success rate improving from 39.8\% to 55.3\%. This demonstrates that MVFF effectively strengthens the multi-view representation rather than merely depending on VGGT’s 3D priors. The improvement is particularly pronounced in geometry-sensitive tasks, such as \textit{Blocks Stack Easy} (from 30\% to 51\%) and \textit{Block Hammer Beat} (from 44\% to 74\%), where accurate multi-view spatial reasoning is crucial.

\noindent\textbf{Effectiveness of AFSC.}  
Finally, we equip the model with the Asynchronous Fast–Slow Collaboration (AFSC) mechanism, which we denoted as R3DP, to inject temporal information on top of VGGT. While the VGGT-only variant introduces 3D priors at each frame, AFSC complements it by modeling temporal dependencies and maintaining feature consistency across time.  

As shown in Table \ref{ablation:all_components}, the integration of AFSC significantly improves task success rates. For example, \textit{Blocks Stack Easy} increases from 51\% to 69\%, and \textit{Block Hammer Beat} improves from 74\% to 77\%. The average performance across all tasks increases from 55.3\% to 62.3\%, demonstrating the effectiveness of the AFSC module in enhancing overall task performance.
Moreover, as shown in Table \ref{frequency_results_transposed}, R3DP substantially reduces inference latency. With R3DP ($\tau=4$), the average inference time of observation encoding across tasks drops to 50.5ms from 78.3ms in DP+VGGT+MVFF, resulting in a significant speedup of 35.5\%. The real-time performance is further improved with $\tau=8$, where the average latency reduces to 40.3ms, further demonstrating the efficiency of our approach in maintaining high accuracy while achieving low-latency control.

Overall, the ablations confirm that R3DP enhances 3D perception, multi-view fusion, and temporal reasoning, while preserving real-time efficiency.

\subsection{Real-world experiments}
\begin{figure}[t]
    \centering

    \begin{minipage}[b]{0.24\linewidth}
        \centering
        \includegraphics[width=\linewidth]{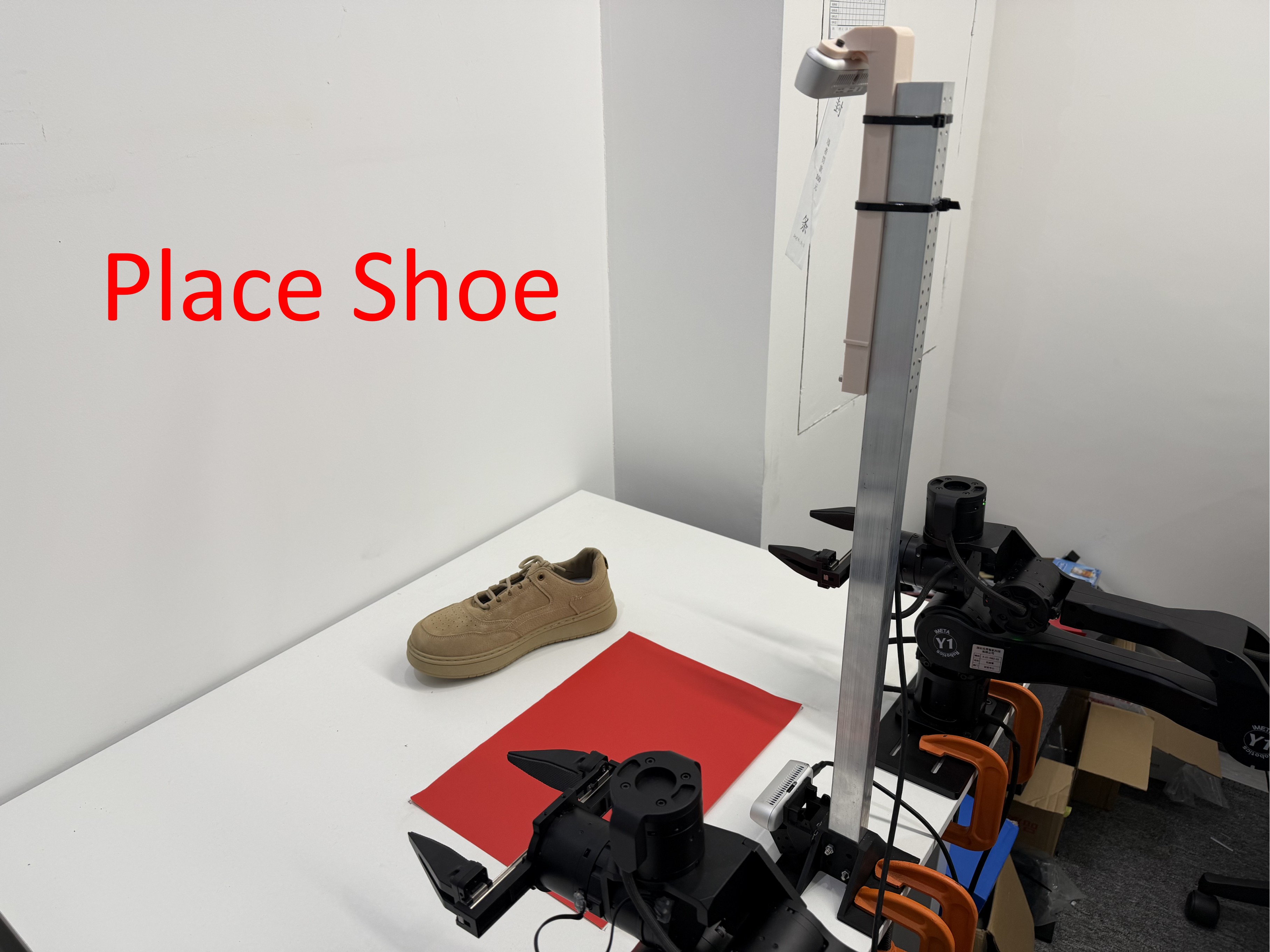} 
    \end{minipage}
    \hfill
    \begin{minipage}[b]{0.24\linewidth}
        \centering
        \includegraphics[width=\linewidth]{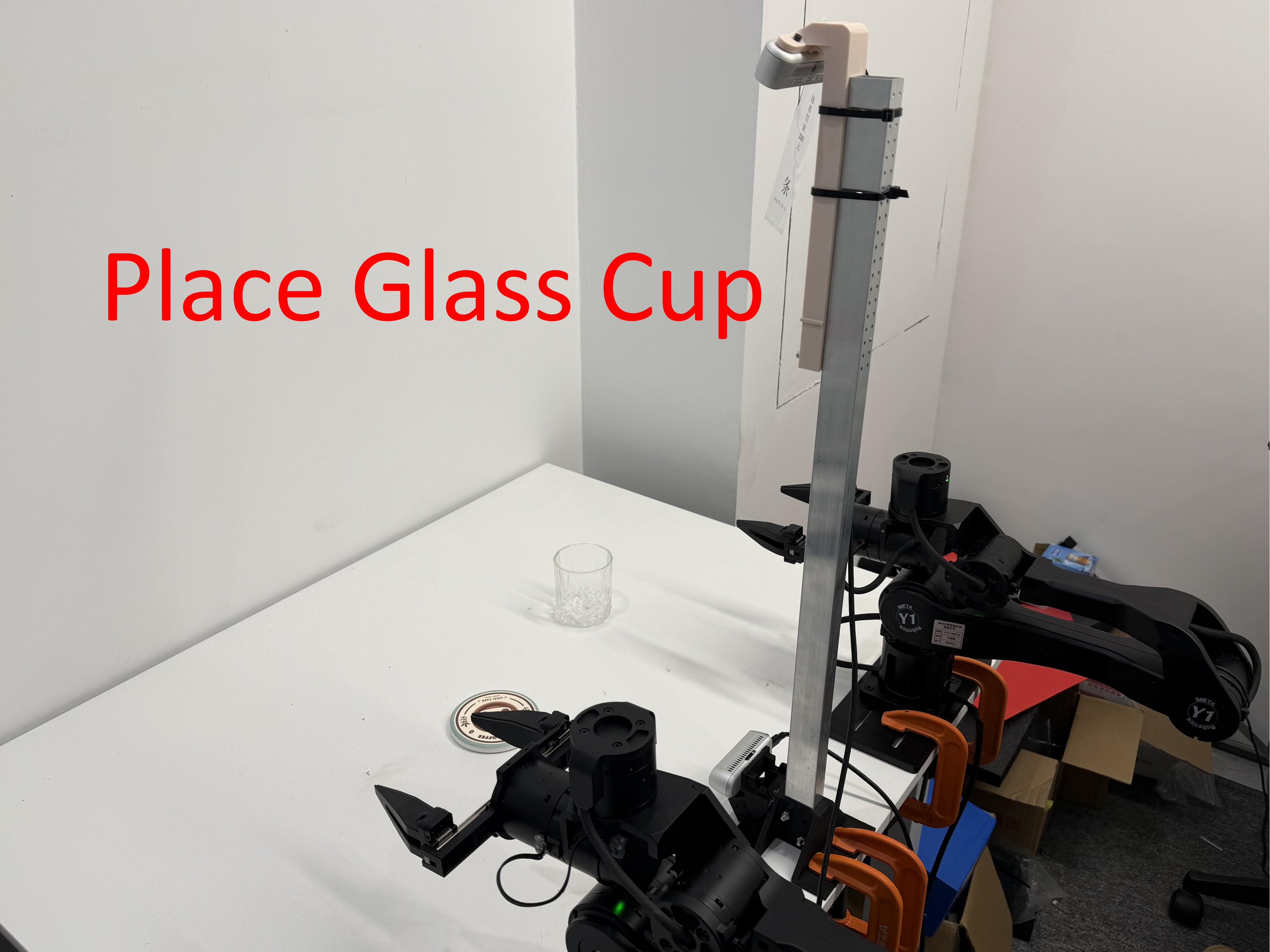} 
    \end{minipage}
    \hfill
    \begin{minipage}[b]{0.24\linewidth}
        \centering
        \includegraphics[width=\linewidth]{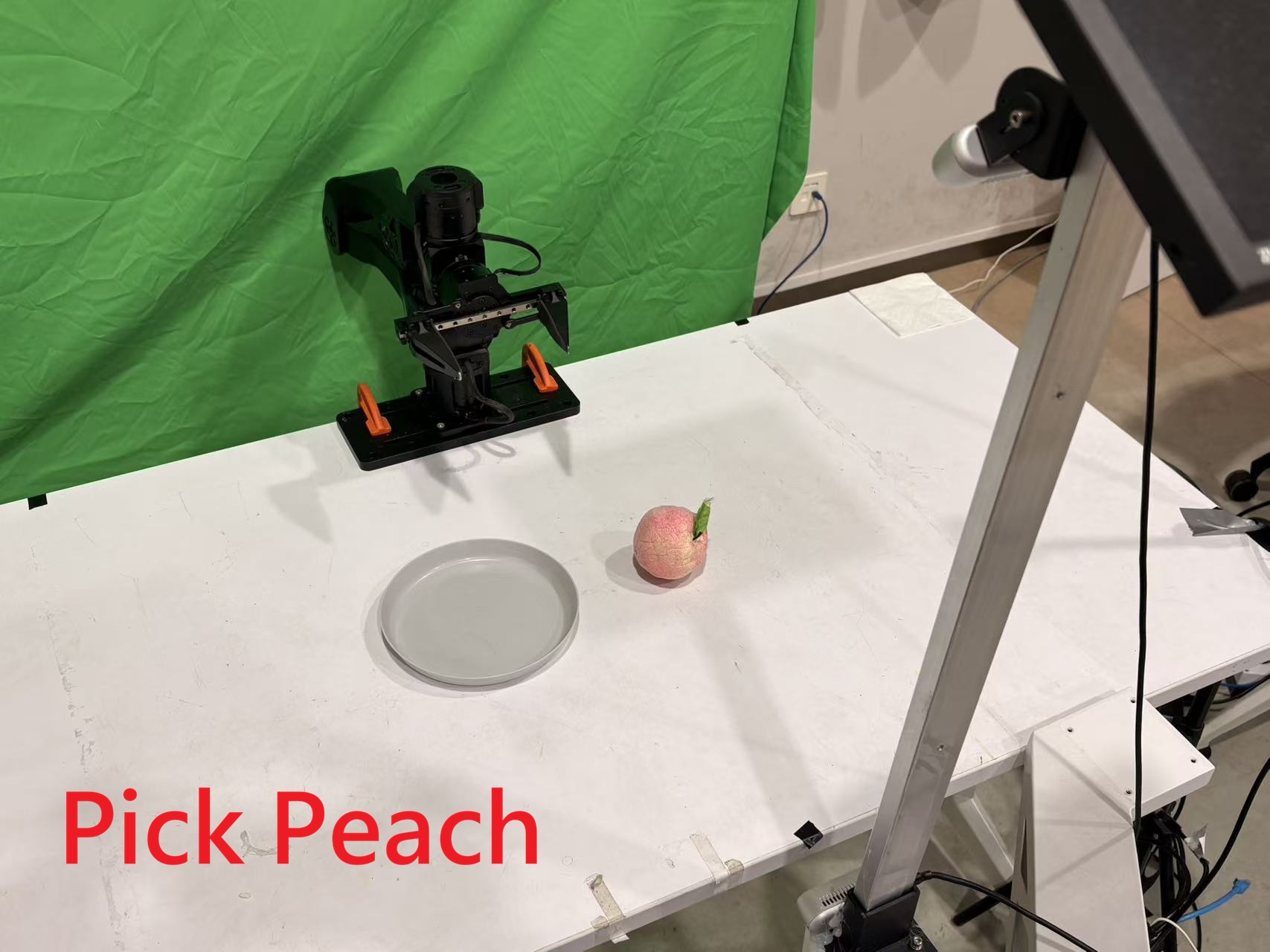} 
    \end{minipage}
    \hfill
    \begin{minipage}[b]{0.24\linewidth}
        \centering
        \includegraphics[width=\linewidth]{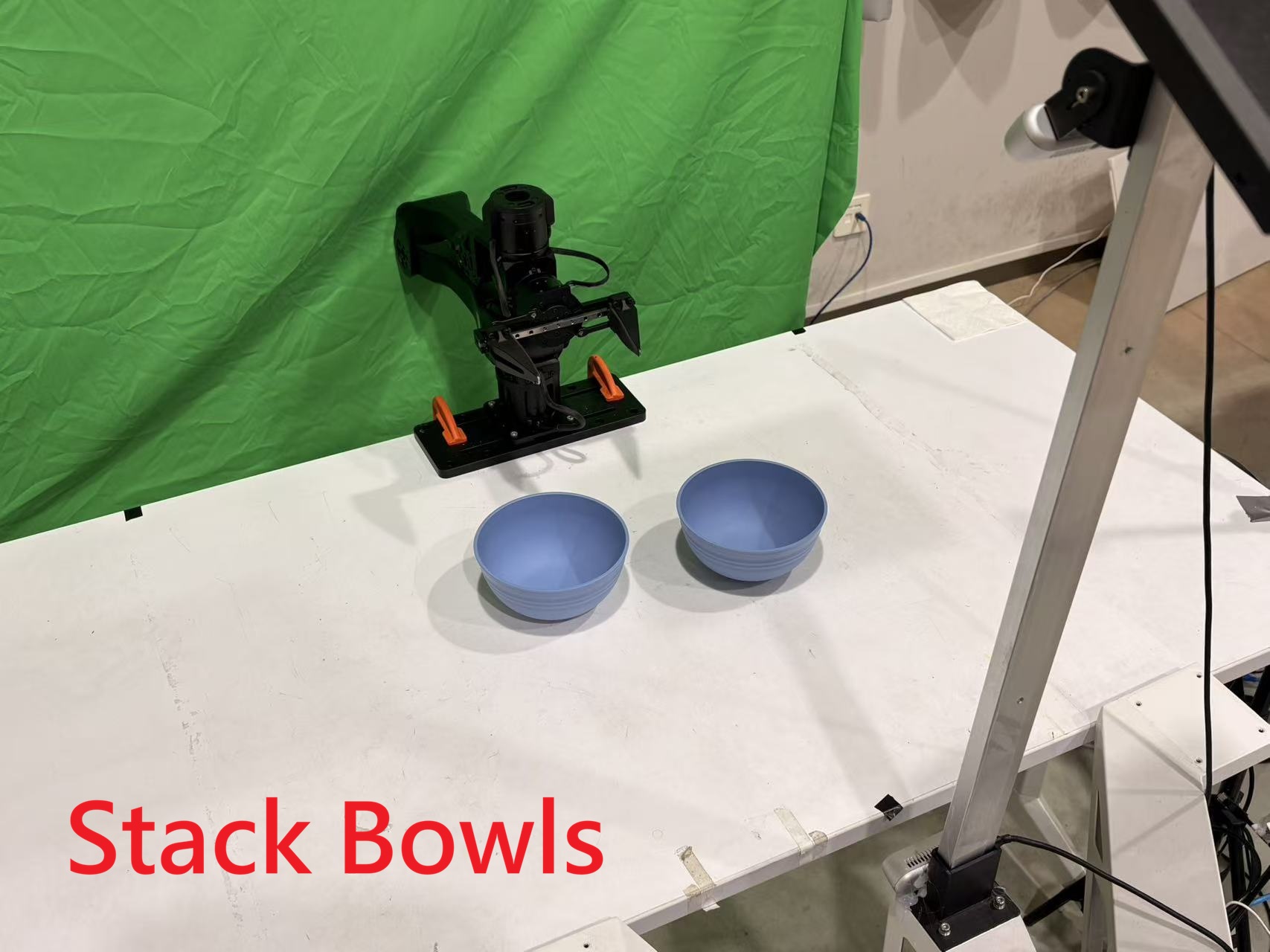} 
    \end{minipage}


    \begin{minipage}[b]{0.24\linewidth}
        \centering
        \includegraphics[width=\linewidth]{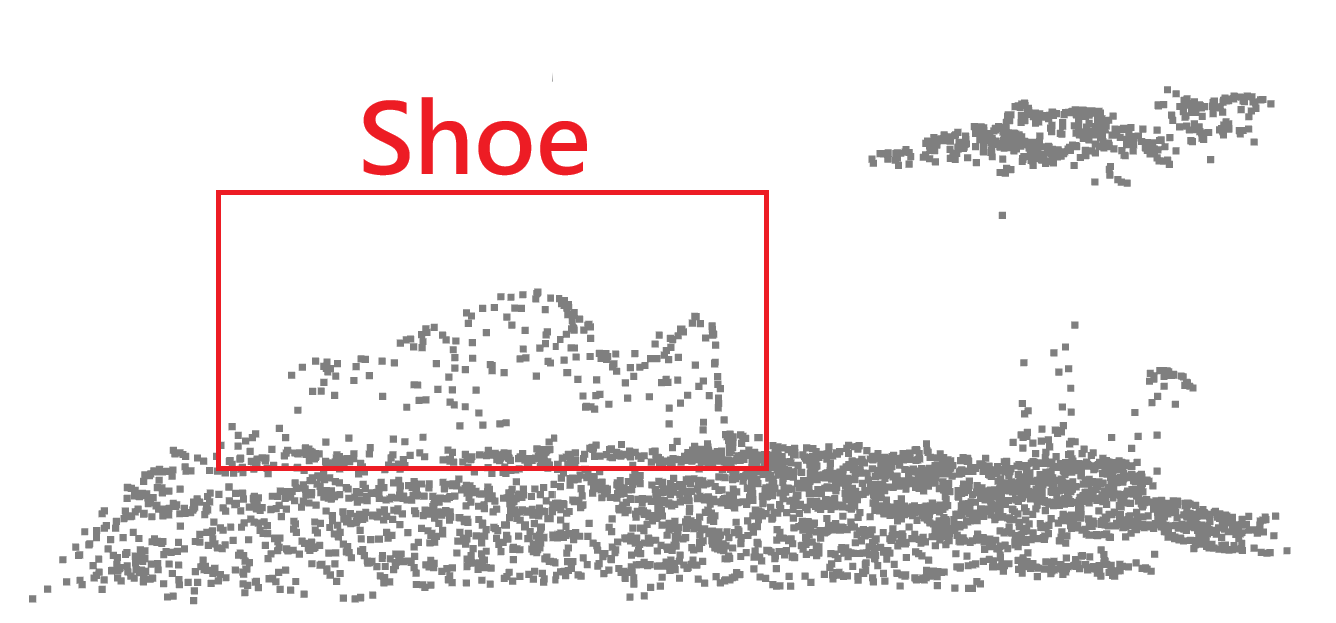} 
    \end{minipage}
    \hfill
    \begin{minipage}[b]{0.24\linewidth}
        \centering
        \includegraphics[width=\linewidth]{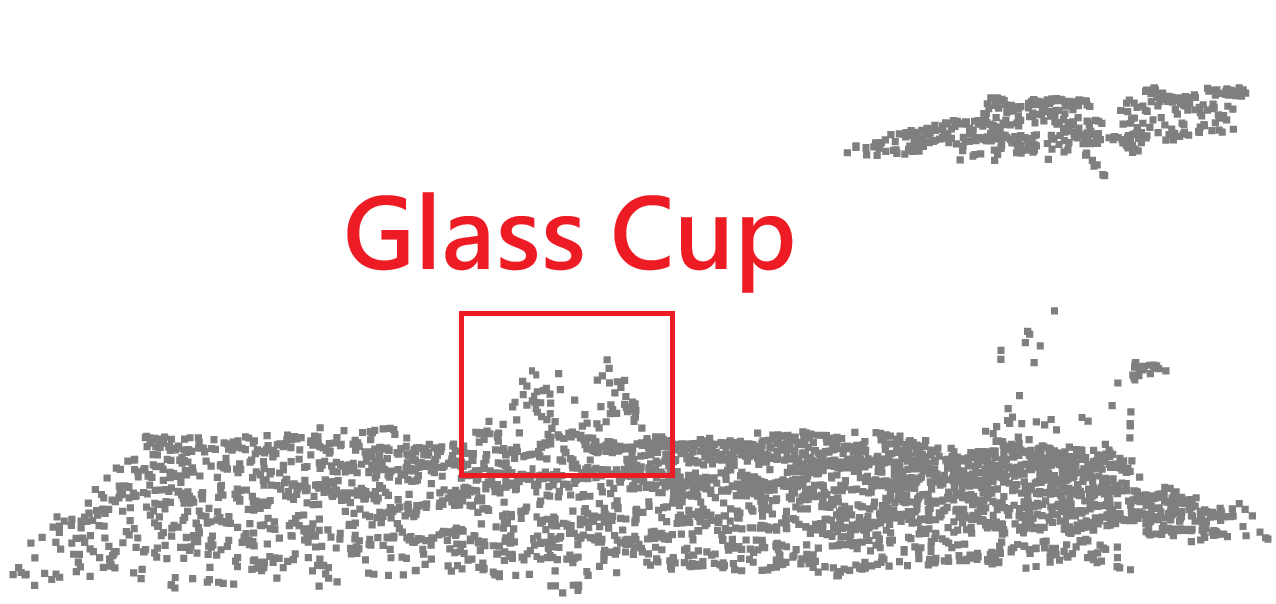} 
    \end{minipage}
    \hfill
    \begin{minipage}[b]{0.24\linewidth}
        \centering
        \includegraphics[width=\linewidth]{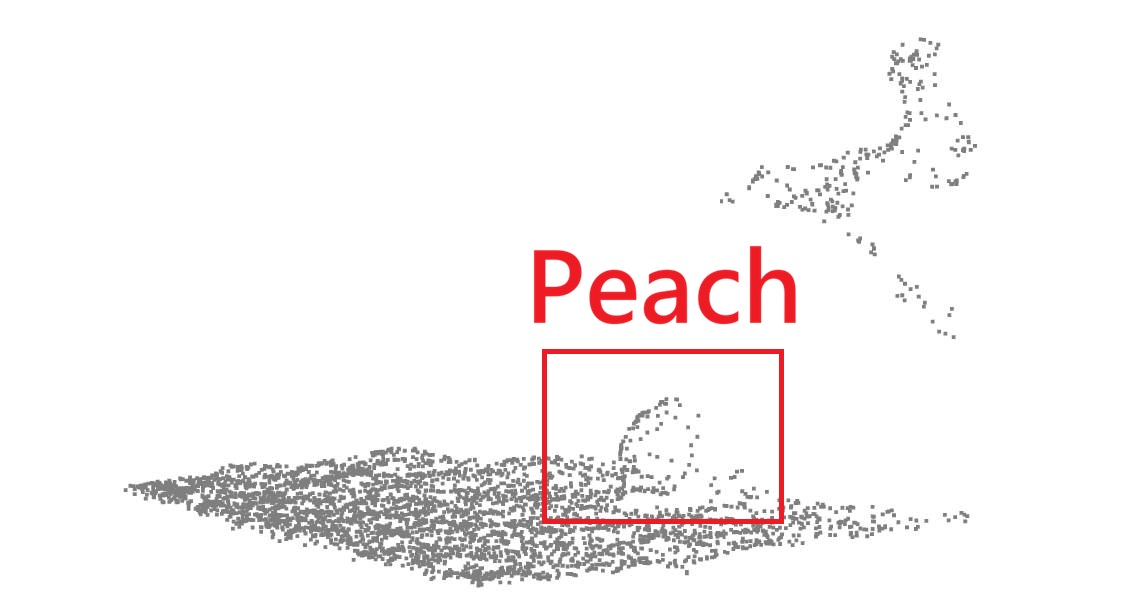} 
    \end{minipage}
    \hfill
    \begin{minipage}[b]{0.24\linewidth}
        \centering
        \includegraphics[width=\linewidth]{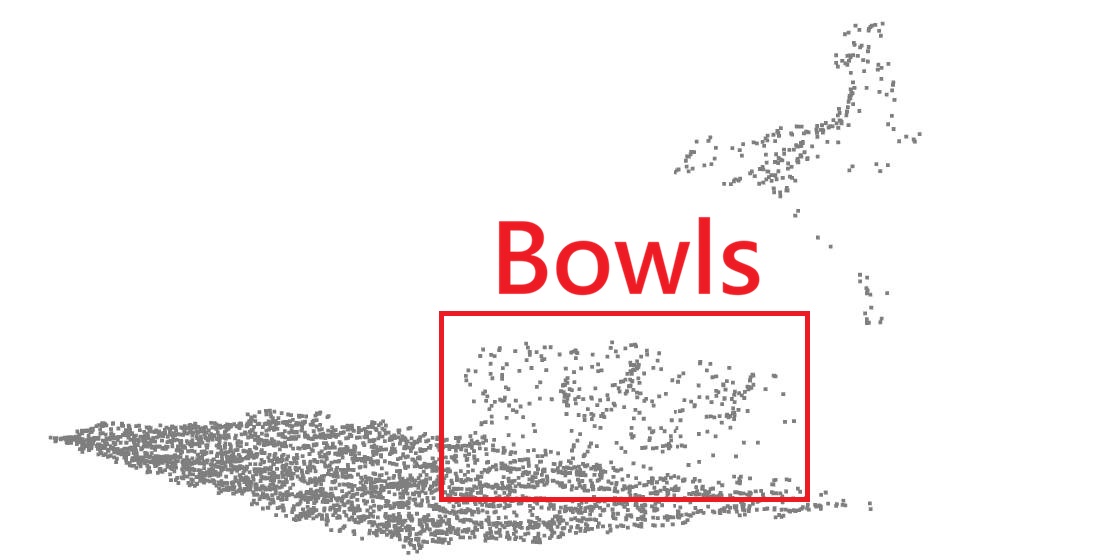} 
    \end{minipage}

    \caption{Real-world experimental platforms and point cloud inputs. We evaluate our method on the ArmBot-Y1 bimanual robot for tasks including Place Shoe and Place Glass Cup, and a single-arm robot for Pick Peach and Stack Bowls. Both platforms are equipped with dual RealSense D435 cameras to generate the ground-truth point clouds used by the DP3 baseline.}
    \label{fig:real-world}
    \vspace{-0.5cm}
\end{figure}

\noindent\textbf{Environments.}
We evaluate the practical applicability and geometric robustness of R3DP on a bimanual robotic system consisting of two ArmBot-Y1 arms and two fixed RealSense D435 cameras.
We select two representative 3D-aware tasks for evaluation: \textit{Place Shoe} and \textit{Place Glass Cup}. These tasks are challenging due to the noisy and incomplete point clouds typically generated by consumer-grade depth sensors (As visualized in Figure~\ref{fig:real-world}), which often degrade the performance of point-cloud-based policies like DP3. Furthermore, to test environmental robustness, we conduct unimanual (single-arm) robot evaluations in a cluttered environment with significant lighting interference, featuring the \textit{Pick Peach} and \textit{Stack Bowls} tasks.

To ensure a fair and rigorous comparison, we train policies on 100 demonstrations for bimanual tasks and 50 for single-arm tasks. All models are validated across 30 independent trials under identical task configurations, with inference performed on a single NVIDIA RTX 4090 to assess success rates and latency.

\begin{table}[t]
    \centering
    \scriptsize
    \renewcommand\arraystretch{1} 
    \tabcolsep=12pt 
    \caption{Real-world performance and inference latency. V and M denote VGGT and MVFF. For model training, we collect 100 episodes for dual-arm tasks and 50 episodes for single-arm tasks, respectively. All models are subsequently evaluated over 30 trials per task to report the success rates. For DP3, point cloud conversion introduces additional computational overhead.}
    \begin{tabular}{l | c c c c}
        \toprule
        \textbf{Task (SR $\uparrow$)} & \textbf{DP} & \textbf{DP3} & \textbf{DP+V+M} & \textbf{R3DP} \\
        \midrule
        Place Shoe      & 46.7\% & 50.0\% & 76.7\% & \textbf{86.7}\% \\
        Place Glass Cup & 23.3\% & 56.7\% & 76.7\% & \textbf{83.3}\% \\
        Pick Peach(Single Arm) & 20\% & 30\% & 46.7\% & \textbf{50}\% \\
        Stack Bowls(Single Arm) & 33.3\% & 56.7\% & \textbf{66.7}\% & \textbf{66.7}\% \\
        \cellcolor{gray!20}\textbf{Average}& \cellcolor{gray!20}30.8\% & \cellcolor{gray!20}48.4\% &\cellcolor{gray!20} 66.7\% &\cellcolor{gray!20} \textbf{71.7\%} \\
        \midrule
        \textbf{Latency (ms $\downarrow$)} & \multicolumn{4}{c}{\textit{Inference Breakdown (RTX 4090)}} \\
        \midrule
        Obs. Encoder    & ---    & 57.24  & 105.1  & \textbf{62.18} \\
        Action Expert   & 67.94  & 58.77  & 66.32  & 69.78 \\
        \bottomrule
    \end{tabular}
    \vspace{-0.5cm}
    \label{tab:real-world}
\end{table}

\noindent\textbf{Implementation Details.} For real-world DP3 deployment, we crop out points corresponding to walls and background from the depth map, downsample it to 4096 points (as visualized in Figure \ref{fig:real-world}),  discarding color information to follow the original setting. All strategies use DDIM as the noise scheduler, with 100 steps during training and 10 steps during denoising.

\noindent\textbf{Results. }As shown in Table \ref{tab:real-world}, R3DP significantly outperforms Diffusion Policy (DP) and DP3 by average margins of 40.9\% and 23.3\%, respectively. While DP3 relies on point clouds, which are often noisy and incomplete in real-world settings (as illustrated in Figure \ref{fig:real-world}), our proposed VGGT and MVFF modules provide superior geometric perception, leading to more stable manipulation.

Furthermore, we analyze the inference efficiency on an NVIDIA RTX 4090 GPU. Although the inclusion of VGGT initially introduces an encoding bottleneck (105.1 ms), our optimized R3DP architecture reduces the observation encoding latency to 62.18 ms—nearly on par with DP3 (57.24 ms) while delivering higher success rates. This balance of efficiency and accuracy demonstrates R3DP’s readiness for real-time, high-performance 3D manipulation.

%% file: R3DP/sec/6_conclusion.tex
\vspace{-0.4cm}
\section{Conclusion}
We presented R3DP, a real-time 3D-aware policy for embodied manipulation that injects strong 3D priors from large 3D vision models into standard image-based imitation policies.
This framework is plug-and-play and can be instantiated on different policy backbones. Experiments show that R3DP substantially improves success rates over multiple baselines while maintaining real-time control.
We believe this offers a promising path toward combining large-scale 3D models with fast control, and plan to extend R3DP to more contact-rich, long-horizon, and reinforcement-learning–driven manipulation.


%% file: R3DP/sec/X_suppl.tex
\clearpage
\renewcommand{\thefigure}{\Alph{figure}}
\setcounter{figure}{0} 

\renewcommand{\thetable}{\Alph{table}}
\setcounter{table}{0} 



\appendix

\section{Simulation Task Description}
We evaluate our method on 10 tasks in the RoboTwin benchmark, 9 of which follow the benchmark’s predefined task suite.  The language description can be found in Table~\ref{tab:task-desc}.

\begin{table*}[!ht]
\vspace{-0.5cm}
\renewcommand{\arraystretch}{1.5}
\centering
\scriptsize
\caption{Descriptions of the RoboTwin manipulation tasks and our additional Tube Insert task. Max Frames denotes the maximum number of frames used to determine whether an evaluation is considered a failure.}
\label{tab:task-desc}
\begin{tabular}{ccc} 
\toprule
\textbf{Task}            & \textbf{Max Frames} & \textbf{Description}   \\ 
\midrule
Bottle Adjust     & 400        & \multicolumn{1}{m{7cm}}{Pick up the bottle on the table headup with the correct arm.}                                                             \\\hline
Block Hammer Beat & 400        & \multicolumn{1}{m{7cm}}{There is a hammer and a block on the table, use the arm to grab the hammer and beat the block. }                      \\ \hline
Shoe Place        & 450        & \multicolumn{1}{m{7cm}}{Use one arm to grab the shoe from the table and place it on the mat.}                                                     \\\hline
Container Place   & 350        & \multicolumn{1}{m{7cm}}{Place the container onto the plate.}                                                                                      \\\hline
Dual Shoes Place  & 600        & \multicolumn{1}{m{7cm}}{Use both arms to pick up the two shoes on the table and put them in the shoebox, with the shoe tip pointing to the left.}\\ \hline
Diverse Bottles Pick & 400        & \multicolumn{1}{m{7cm}}{Pick up one bottle with one arm, and pick up another bottle with the other arm.}                                   \\\hline
Block Handover       & 600        & \multicolumn{1}{m{7cm}}{Use the left arm to grasp the red block on the table, handover it to the right arm and place it on the blue pad.}  \\\hline
Put Apple Cabinet    & 600        & \multicolumn{1}{m{7cm}}{Use arm movement to open cabinet and grab apple, then place apple inside and close cabinet.}                    \\\hline
Blocks Stack Easy    & 600        & \multicolumn{1}{m{7cm}}{Place the black block on top of the red block at the center.}                                                      \\\hline
Tube Insert          & 500        & \multicolumn{1}{m{7cm}}{Pick up tube and place it into container using arm movement.}                                                      \\
\bottomrule
\vspace{-1.5cm}
\end{tabular}
\end{table*}

\subsection{Tube Insert Task Details}
\begin{figure}[h]
\centering
\includegraphics[width=0.45\columnwidth]{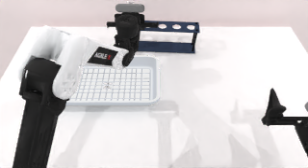}
\includegraphics[width=0.45\columnwidth]{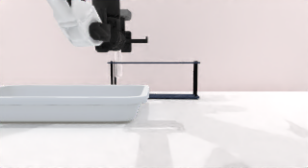}
\caption{The head and front camera view of Tube Insert task.}
\label{fig:tubegrasp}
\end{figure}

In order to specifically assess the capability of R3DP in manipulating transparent objects, we introduce a customized simulated task  Tube Insert in addition to the pre-defined tasks in RoboTwin. A visualization of \textit{Tube Insert} in simulation can be found in Fig.~\ref{fig:tubegrasp}.

\noindent\textbf{Task Setup.}  The background and lighting conditions of the \textit{Tube Insert} task are consistent with those defined in the RoboTwin benchmark. Three custom objects are placed on the tabletop:  a test-tube rack, a white rectangular plastic container with a grid pattern, and a transparent glass tube positioned inside the container.

\noindent\textbf{Example Procedure.} 1) The gripper first moves to a position 10\,cm above the tube.  
2) It approaches the tube and lifts it vertically by 4\,cm.  
3) The gripper then retracts and rotates to an upright orientation, ensuring that the tube opening faces upward.  
4) Next, it moves to a position 20\,cm above the target slot on the test-tube rack.  
5) The gripper descends, inserts the tube into the rack, and releases it. 

\noindent\textbf{Success Criterion.} Within 500 frames, the tube must be transported such that its final position is within 5\,cm of the designated target location inside the test-tube rack along all three spatial axes.

\section{Real-world Task Description}

\subsection{Experiment Environment Settings}
\begin{figure}[!ht]
    \centering

    \begin{subfigure}[b]{0.495\linewidth}
        \centering
        \includegraphics[width=0.49\linewidth]{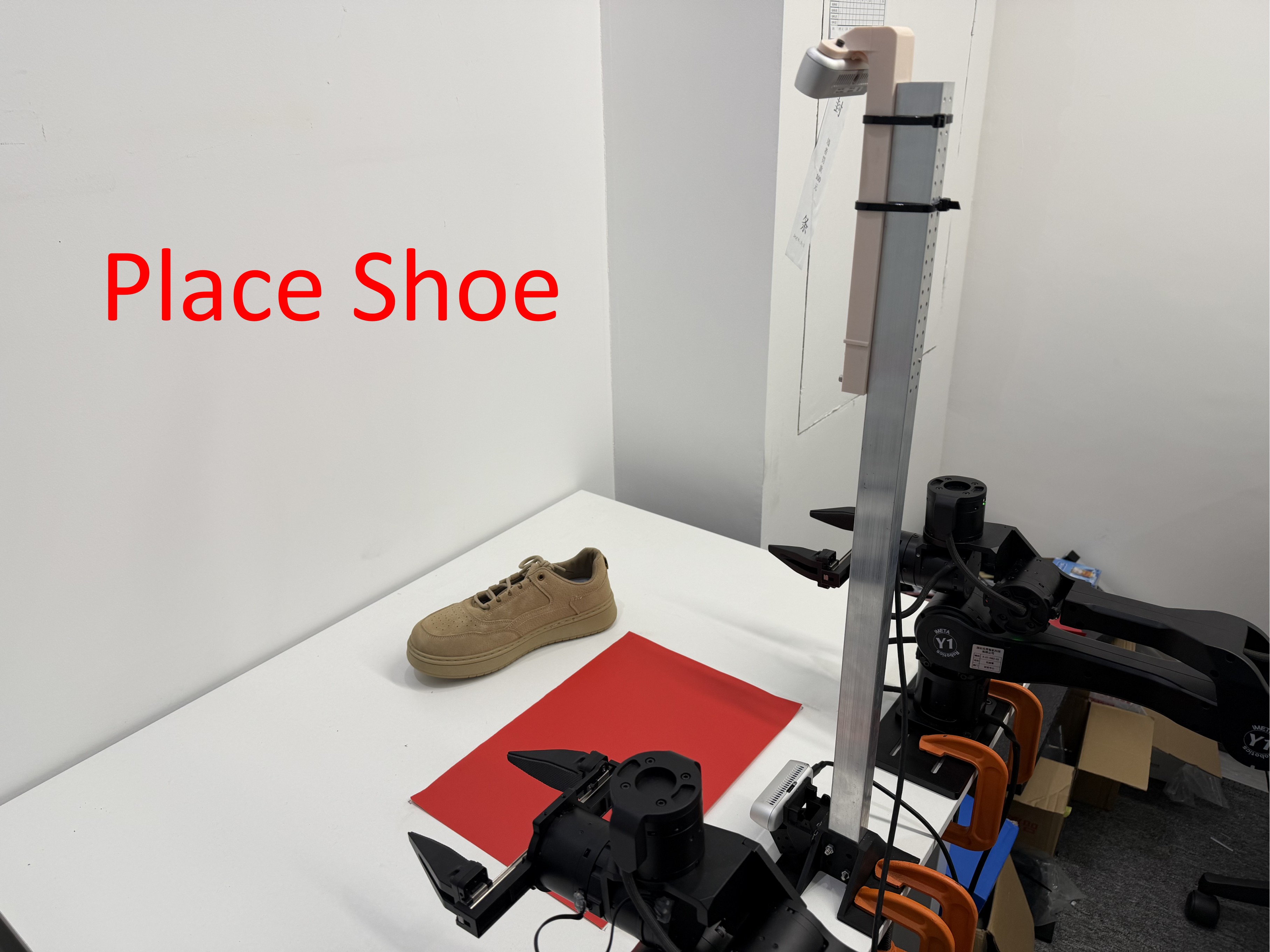}
        \hfill
        \includegraphics[width=0.49\linewidth]{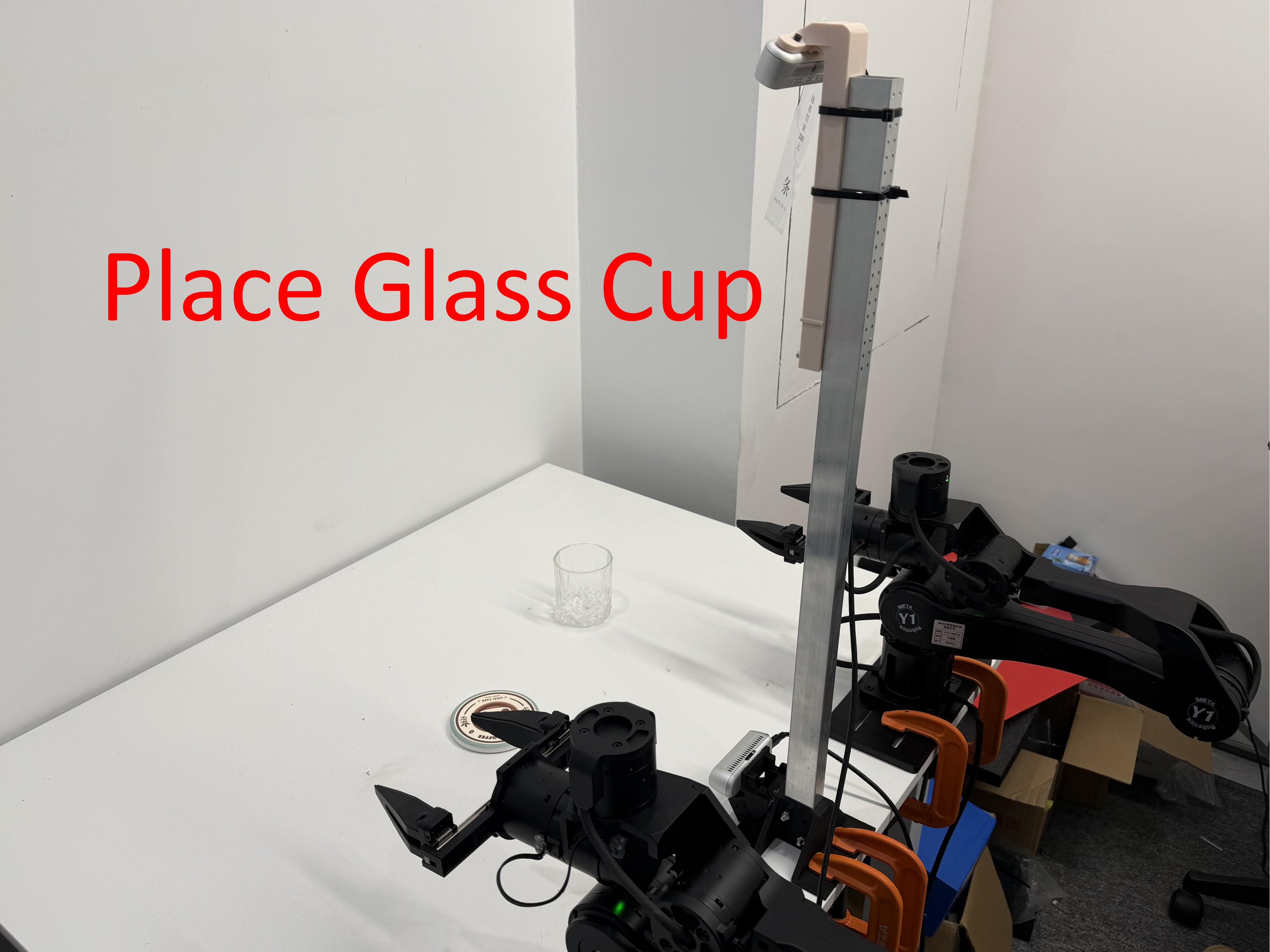}
    \end{subfigure}
    \hfill
    \begin{subfigure}[b]{0.495\linewidth}
        \centering
        \includegraphics[width=0.49\linewidth]{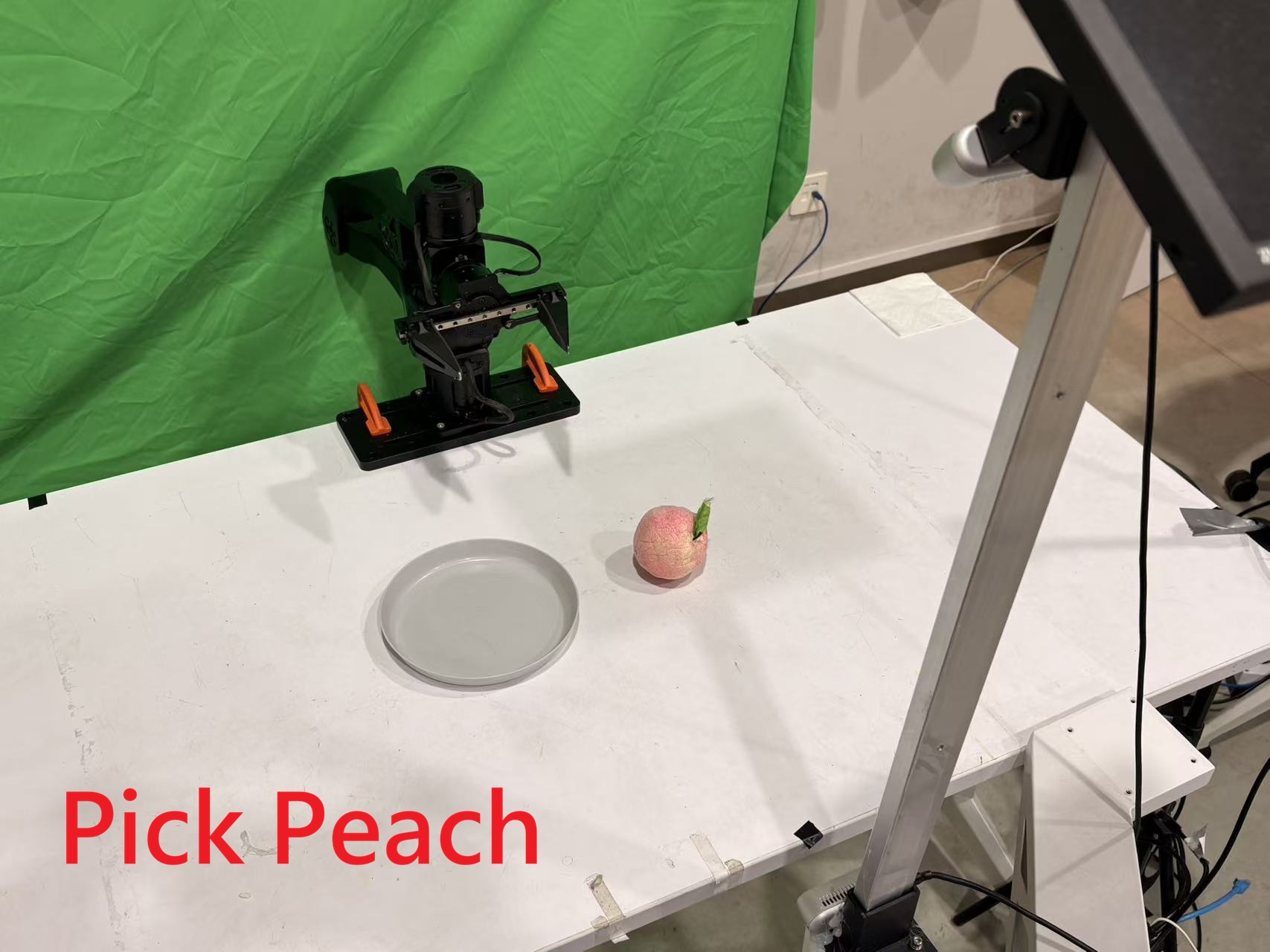}
        \hfill
        \includegraphics[width=0.49\linewidth]{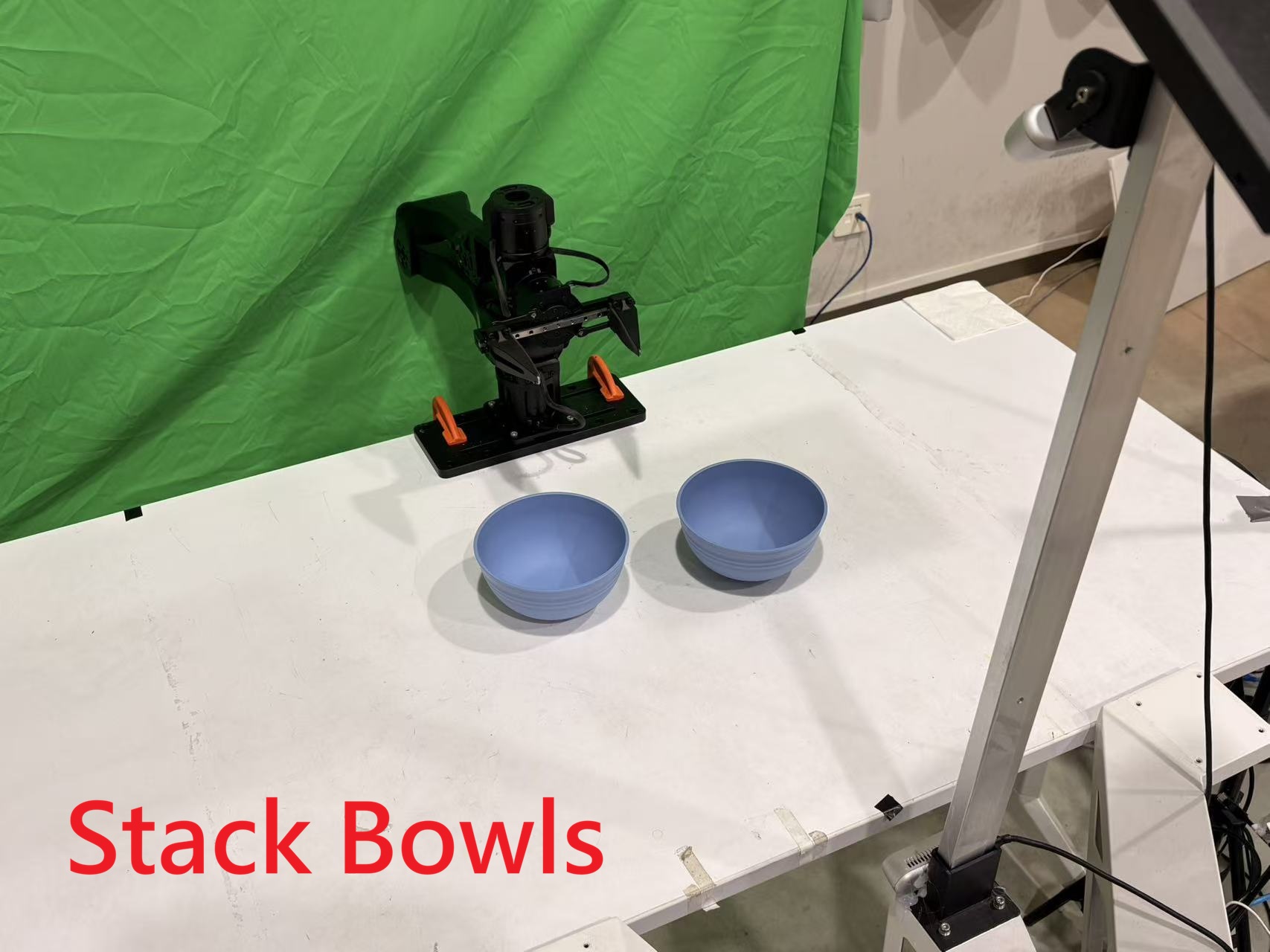}
    \end{subfigure}
    \vspace{-0.5cm}
    \caption{Real-world experiment settings and tasks. The first two images correspond to Setting A, while the last two represent Setting B.}
    \label{fig:real-world-supp}
\end{figure}

To evaluate the robustness of R3DP, we conducted real-robot experiments across two distinct experimental setups (\cref{fig:real-world-supp}):
\begin{itemize}
    \item Setting A (Controlled Indoor Environment): This setting is in a windowless, confined room. Fully controllable illumination. Clean white wall and tabletop. Two Armbot-Y1 robotic arms are positioned in parallel 60 cm from each other. Visual inputs are captured by two Intel RealSense D435 cameras, designated the front-view and head-view cameras, respectively. Specifically, the front-view camera is centered between the bases of the two manipulators. The head-view camera is mounted 
    80 cm directly above the front-view camera, maintaining a 60-degree downward tilt.
    \item Setting B (Open Laboratory Environment): This setting is in a wide laboratory. Subject to ambient natural light and semi-controlled artificial lighting. Clean white tabletop against a green-screen background. A single Armbot-Y1 arm is employed for these tasks. While the relative spatial transformation  between the two cameras remains identical to Setting A, the camera rig is repositioned 
    90 cm in front of the robotic arm. In this configuration, the cameras face the manipulator instead of being co-directional with the end-effector to mitigate self-occlusion during single-arm manipulation.
\end{itemize}

\subsection{Data Collection} \label{sec:real_data_collection}

\begin{figure}[t]
    \centering

    \begin{subfigure}[b]{0.24\linewidth}
        \centering
        \includegraphics[width=\linewidth]{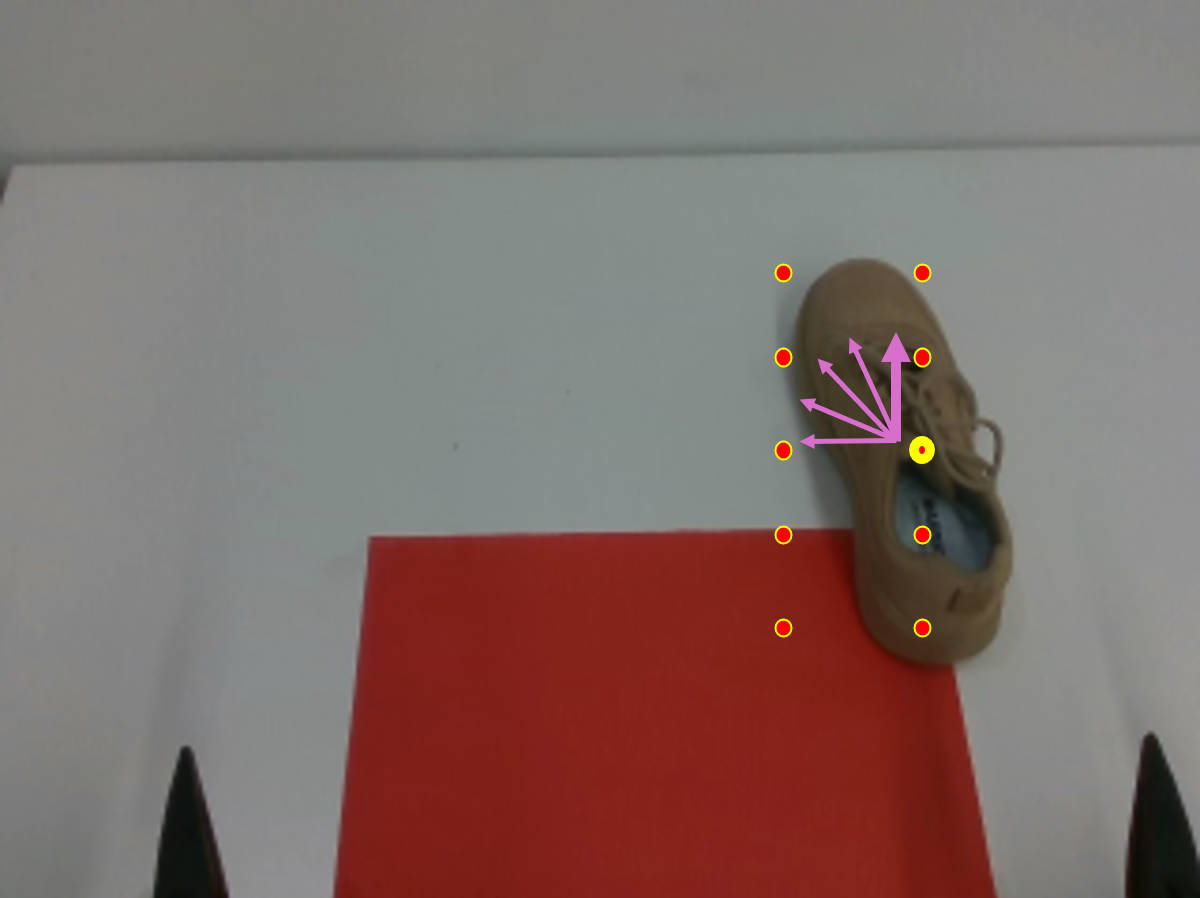}
        \caption{Place Shoe}
    \end{subfigure}
    \hfill
    \begin{subfigure}[b]{0.24\linewidth}
        \centering
        \includegraphics[width=\linewidth]{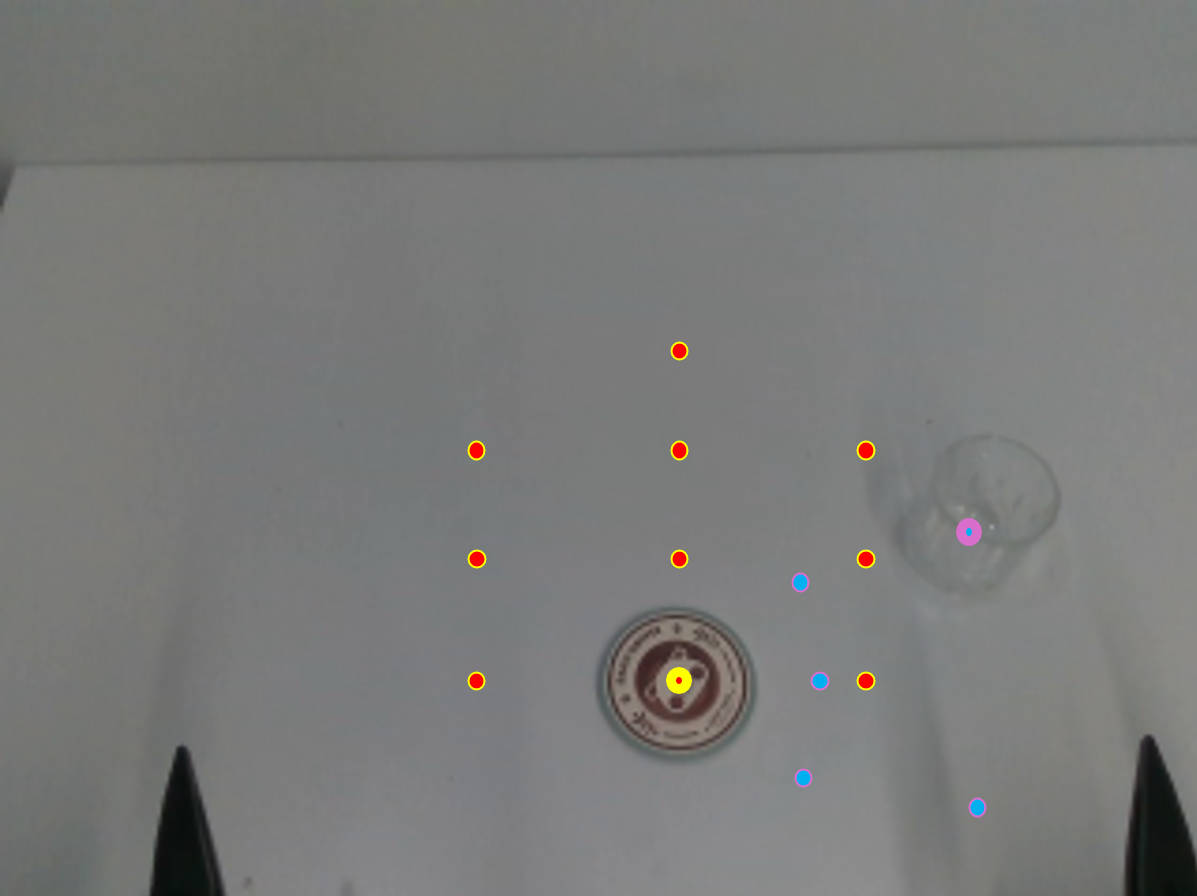}
        \caption{Place Glass Cup}
    \end{subfigure}
    \hfill
    \begin{subfigure}[b]{0.24\linewidth}
        \centering
        \includegraphics[width=\linewidth]{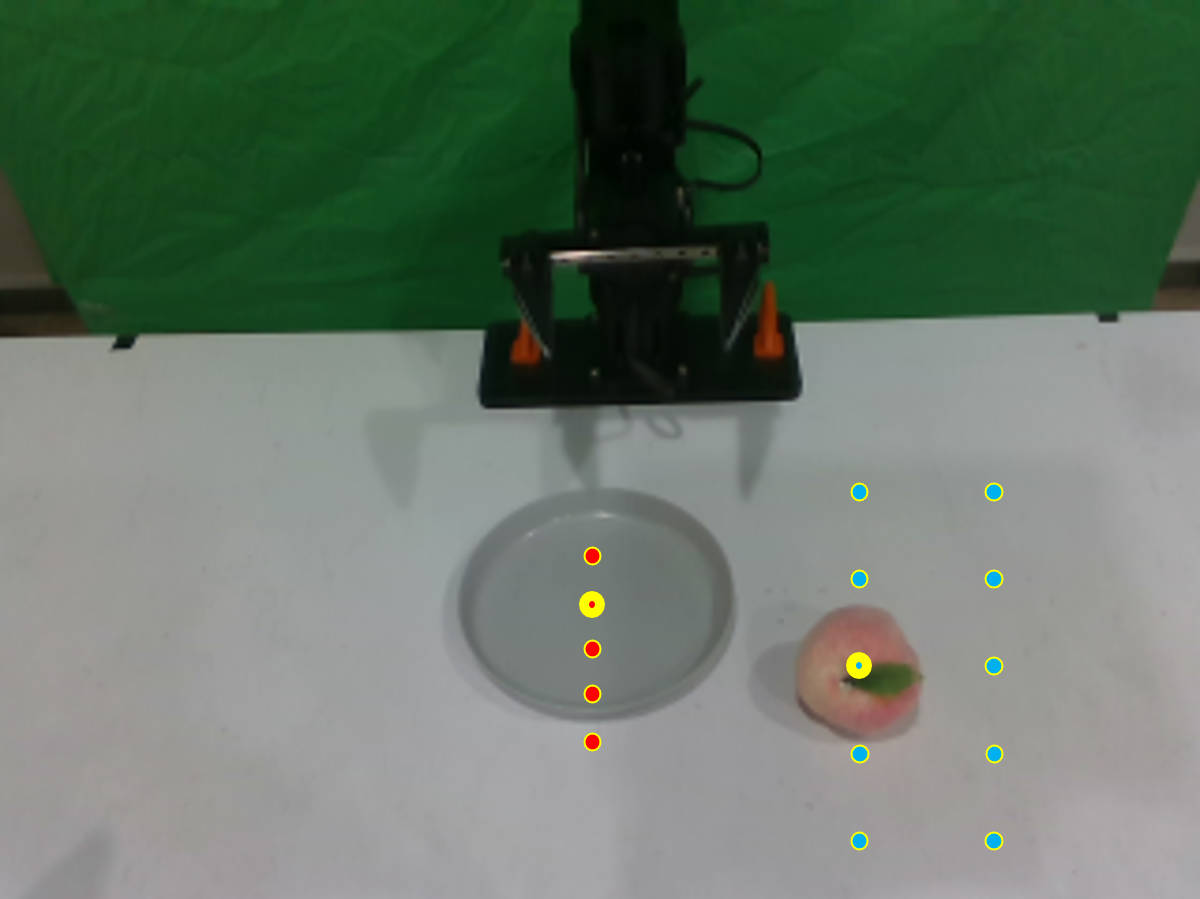}
        \caption{Pick Peach}
    \end{subfigure}
    \hfill
    \begin{subfigure}[b]{0.24\linewidth}
        \centering
        \includegraphics[width=\linewidth]{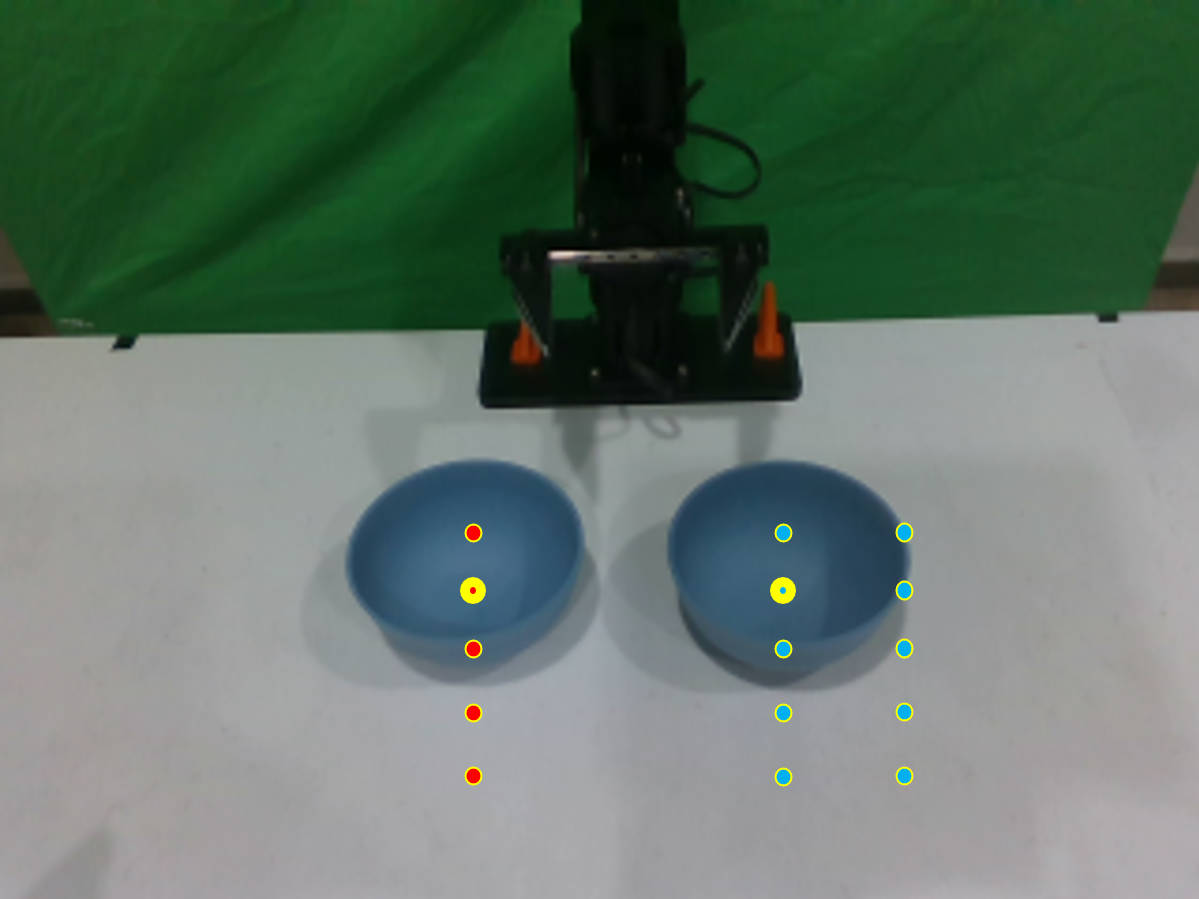}
        \caption{Stack Bowls}
    \end{subfigure}

    \caption{Real-world experiment data collection settings. Different colors represent different objects. The yellow edge means absolute position, and magenta means relative position or direction. The highlighted points represent the selected placement positions of the objects in the sample images.}
    \label{fig:real-world-data}
\end{figure}

As shown in~\cref{fig:real-world-data}, we designated specific positions for data collection based on the head view. All positions were determined from head-view images without precise metric calibration.
\begin{itemize}
    \item \textit{Place Shoe:} We selected ten positions for the midpoint of the shoe. For each position, the direction of the shoe tip was uniformly divided into five intervals, ranging from the front to a direction perpendicular to the front and facing the centerline. Symmetric configurations were handled analogously. In all training data, the gripper was constrained to grasp the outer side of the shoe.
    \item \textit{Place Glass Cup:} We selected ten positions for the midpoint of the coaster. For each coaster position, the cup was placed at five distinct relative locations: three closer and two farther. Symmetric configurations were handled analogously. In all training data, the gripper was constrained to grasp the outer side of the cup.
    \item \textit{Pick Peach:} We selected five positions for the plate and ten positions for the peach. In all training data, the peach leaf was oriented toward the camera to ensure visual consistency.
    \item \textit{Stack Bowls:} We selected five positions for the left bowl (fixed) and ten positions for the right bowl (to be moved). In all training data, the gripper was constrained to grasp only the right side of the bowl according to the camera view.
\end{itemize}





\subsection{Detailed Evaluation Results}
During our evaluation, we observed that different models fail in different cases. As shown in Table \ref{tab:failure_analysis}, instead of a binary success/failure metric, we introduced a fine-grained categorization of failure modes to discuss the performance of different models. Based on the categorization, we listed detailed evaluation results in Table \ref{tab:detailed_results}.

\begin{table}[t]
\setlength{\tabcolsep}{4pt}
\centering
\scriptsize
\caption{Detailed categorization of failure modes across four real-world tasks.}
\label{tab:failure_analysis}
\begin{tabular}{lll}
\toprule
\textbf{Task} & \textbf{Notation} & \textbf{Description} \\
\midrule
\multirow{4}{*}{Place Shoe} & Success & 1 - Successfully placed the shoe inside the red area horizontally. \\
\cmidrule{2-3}
    & \multirow{2}{*}{Failure} & 0 - Failed to grasp. \\
    &        & 2 - Placed the shoe out of the area or in incorrect orientation. \\
\midrule
\multirow{4}{*}{Place Glass Cup} & Success & 1 - Successfully placed glass cup fully onto the coaster. \\
\cmidrule{2-3}
    & \multirow{2}{*}{Failure} & 0 - Failed to grasp. \\
    &        & 2 - Placed the glass cup out of the coaster. \\
\midrule
\multirow{4}{*}{Pick Peach} & Success & 1 - Successfully placed the peach onto the plate. \\
\cmidrule{2-3}
    & \multirow{3}{*}{Failure} & 0 - Attempted to grasp at the wrong position. \\
    &        & 2 - Approached the correct position but failed to grasp. \\
    &        & 3 - Placed the peach out of the plate. \\
\midrule
\multirow{4}{*}{Stack Bowls} & Success & 1 - Successfully stacked the right bowl onto the left one. \\
\cmidrule{2-3}
    & \multirow{3}{*}{Failure} & 0 - Attempted to grasp at the wrong position. \\
    &        & 2 - Approached the correct position but failed to grasp. \\
    &        & 3 - Placed the right bowl out of the left bowl or knocked it over. \\
\bottomrule
\end{tabular}
\end{table}

\begin{table}[t]
\setlength{\tabcolsep}{4pt}
\centering
\scriptsize
\caption{Detailed real-world evaluation results}
\begin{tabular}{ccccc}
\toprule
Task                & DP                                   & DP3                                  & DP+VGGT                              & R3DP                                 \\
\midrule
Place Shoe      & \makecell{1012011100\\1110112000\\1000011210} & \makecell{0100100110\\1100112101\\1011020211} & \makecell{1111102121\\1011101110\\1111111110} & \makecell{1111111111\\1112111111\\1221111211} \\
\midrule
Place Glass Cup & \makecell{1010000011\\0000000001\\2010010020} & \makecell{1000010100\\1111010111\\1001011011} & \makecell{1110111110\\1111000111\\0111110111} & \makecell{1110111111\\1112111112\\1121110111} \\
\midrule
Pick Peach      & \makecell{0022212211\\0203010021\\0220013022} & \makecell{1000110110\\2000002001\\0001021010} & \makecell{1100211100\\2121131210\\1100120010} &                   \makecell{0112210110\\2211111201\\1101200100}                   \\
\midrule
Stack Bowls     & \makecell{1221030321\\3221123231\\2212211212} & \makecell{3113111001\\1313031311\\1133113011} & \makecell{1001133111\\1311021311\\2111131111} & 
\makecell{1100111333\\0111131131\\1211111112} \\      
\bottomrule
\end{tabular}
\label{tab:detailed_results}
\end{table}

\section{Full Experiments on Different $\tau$}
In our framework, $\tau$ is a hyperparameter that controls the interval between two queries to the slow branch. 
In the main paper, we report results with $\tau$ = 4 and $\tau$ = 8. 
In Table \ref{tab:tau}, we provide the complete experimental results under a wider range of $\tau$ values, including $\tau$ = 2, 4, 8, 16, and 32.
To assess the sensitivity of our framework to this hyperparameter, we report the average success rate and the corresponding standard deviation across all tasks for each $\tau$ configuration.

\vspace{-0.5cm}
\begin{table}[!ht]
\centering
\scriptsize
\caption{Performance under different values of the hyperparameter $\tau$.}
\label{tab:tau}
\setlength{\tabcolsep}{4pt}
\begin{tabular}{lccccc|c}
\toprule
Task & $\tau=2$ & $\tau=4$ & $\tau=8$ & $\tau=16$ & $\tau=32$ & {Average} \\ 
\midrule
Block Hammer Beat & 60\% & 77\% & 77\% & \textbf{87}\% & 85\% & 77.2\% $\pm$ 9.52\% \\ 
Block Handover & 75\% & \textbf{95}\% & 93\% & 93\% & 90\% & 89.2\% $\pm$ 7.28\% \\
Bottle Adjust & 56\% & \textbf{62}\% & 56\% & 53\% & 53\% & 56.0\% $\pm$ 3.29\% \\
Blocks Stack Easy & 64\% & 69\% & \textbf{72}\% & 67\% & 69\% & 68.2\% $\pm$ 2.64\% \\
Shoe Place  & 73\% & 72\% & 68\% & \textbf{74}\% & 61\% & 69.6\% $\pm$ 4.76\% \\
Container Place & 50\% & \textbf{63}\% & 54\% & 57\% & 53\% & 55.4\% $\pm$ 4.41\% \\
Dual Shoes Place & 20\% & \textbf{24}\% & 20\% & 21\% & 19\% & 20.8\% $\pm$ 1.72\% \\
Diverse Bottles Pick & 28\% & 31\% & 32\% & \textbf{34}\% & 33\% & 31.6\% $\pm$ 2.06\% \\
Put Apple Cabinet & \textbf{100}\% & \textbf{100}\% & 99\% & \textbf{100}\% & \textbf{100}\% & 99.8\% $\pm$ 0.40\% \\
Tube Insert & 90\% & 97\% & 97\% & 98\% & \textbf{99}\% & 98.0\% $\pm$ 0.89\% \\
\bottomrule
\end{tabular}
\end{table}

\begin{table}[!ht]
\captionof{table}{\footnotesize Inference latency for each $\tau$.}
\vspace{-0.25cm}
\centering
\scriptsize
\setlength{\tabcolsep}{8pt}
\begin{tabular}{l | c c c c c}
\toprule
\textbf{Latency (ms)} & \textbf{$\tau$=2} & \textbf{$\tau$=4} & \textbf{$\tau$=8} & \textbf{$\tau$=16} & \textbf{$\tau$=32} \\
\midrule
Obs. Encoder  & 66.17 & 50.52 & 40.27 & 33.31 & \textbf{31.04} \\
Action Expert & 54.35 & 55.27 & 56.66 & 51.83 & 52.25 \\
\bottomrule
\end{tabular}
\vspace{-9pt}
\label{tab:tau_latency}
\end{table}

\section{Implementation Details}
In this section, we provide additional implementation details of R3DP used in all our experiments. We first outline the overall training pipeline, highlighting the two-stage design that decouples 3D feature distillation from policy learning. We then describe the pre-training setup of the Temporal Feature Prediction Network (TFPNet) and the training details of R3DP for both RoboTwin simulation and real-world experiments respectively, covering their training paradigms, data sampling strategies, and optimization hyperparameters. The training protocols and hyperparameters are the same for both simulation and real-world experiments. The difference lies only in the training data source and evaluation environment.

\subsection{Overall Training Pipeline}
Our training follows a two-stage pipeline. In the first stage, we pre-train the Temporal Feature Prediction Network (TFPNet) using supervision from a frozen VGGT backbone, learning to predict 3D-aware features across multi-view manipulation trajectories. In the second stage, we integrate TFPNet into the AFSC module for policy training, where VGGT and TFPNet process frames at different temporal rates. The R3DP policy is then trained end-to-end, with both VGGT and TFPNet frozen during this phase.

\subsection{TFPNet Pre-training Details}
The TFPNet is trained independently for each task using a sequence-based learning approach. For RoboTwin simulation, the dataset consists of approximately 200,000 sequences per task from the same 100 trajectories used for imitation learning, with each sequence containing 4 consecutive frames captured from two different camera views, head and front. For real-world experiments, we use the same model architecture and training protocol, while the training data are constructed from the collected real-world demonstrations described in Sec. \ref{sec:real_data_collection}.For each task, we train a separate model using a dataset constructed solely from the sequences of that task.d


During pre-training, we freeze the VGGT encoder, and only optimize the parameters of the TFPNet. For each sequence, we extract intermediate VGGT tokens as supervision targets and let TFPNet iteratively predict the corresponding features over time, starting from the ground-truth features of the first frame. The training loss is given by a sequence feature-matching objective, while depth reconstruction is used for validation and optional visualization but is not included in the training objective.

The full set of optimization and data-loading hyperparameters is summarized in Table~\ref{tab:tfpnet-hparams}.

\begin{table*}[htbp]
\centering

\begin{minipage}[c]{0.46\textwidth}
\centering
\footnotesize

\captionsetup{font=scriptsize}
\captionof{table}{TFPNet training hyperparameters.}
\vspace{-0.2cm}
\renewcommand\arraystretch{1.35}
\resizebox{\linewidth}{!}{%
\begin{tabular}{ll}
\toprule
\textbf{Parameter} & \textbf{Value} \\
\midrule
sequence\_length              & 4 \\
num\_views                    & 2 (head, front) \\
image\_size                   & $308 \times 168$ \\
batch\_size                   & 12 \\
accumulation\_steps           & 1 \\
optimizer.lr                  & $1.0 \times 10^{-4}$ \\
optimizer.betas               & $(0.9, 0.999)$ \\
optimizer.eps                 & $1.0 \times 10^{-8}$ \\
optimizer.weight\_decay       & $1.0 \times 10^{-4}$ \\
num\_epochs                   & 100 \\
gradient\_clip\_norm          & 1.0 \\
mixed\_precision\_dtype       & bfloat16 \\
gradient\_accumulation\_steps & 1 \\
gpu                           & 8$\times$NVIDIA A800 \\
per\_task\_sequences          & 200,000 \\
\bottomrule
\end{tabular}
}
\label{tab:tfpnet-hparams}



\end{minipage}
\hfill
\begin{minipage}[c]{0.50\textwidth}
\vspace{0pt}
\captionsetup{font=scriptsize}
\captionof{table}{R3DP Training hyperparameters.}
\vspace{-0.2cm}
\centering
\footnotesize

\resizebox{\linewidth}{!}{%
\begin{tabular}{ll}
\toprule
\textbf{Parameter} & \textbf{Value} \\
\midrule
horizon                            & 8 \\
n\_obs\_steps                      & 1 \\
n\_action\_steps                   & 8 \\
n\_latency\_steps                  & 0 \\
num\_inference\_steps              & 100 \\
obs\_as\_global\_cond              & True \\
dataloader.batch\_size             & 8 \\
dataloader.num\_workers            & 0 \\
dataloader.shuffle                 & True \\
dataloader.pin\_memory             & True \\
dataloader.persistent\_workers     & False \\
optimizer.\_target\_               & torch.optim.AdamW \\
optimizer.lr                       & $2.0\times10^{-5}$ \\
optimizer.betas                    & $(0.95, 0.999)$ \\
optimizer.eps                      & $1.0\times10^{-8}$ \\
optimizer.weight\_decay            & $1.0\times10^{-6}$ \\
gradient\_clip\_norm               & 1.0 \\
training.lr\_scheduler             & cosine \\
training.lr\_warmup\_steps         & 500 \\
training.num\_epochs               & 600 \\
training.gradient\_accumulate\_every & 1 \\
training.use\_ema                  & True \\
training.max\_grad\_norm           & 1.0 \\
gpu                                & 8$\times$NVIDIA RTX 4090 \\
\bottomrule
\end{tabular}
}
\label{tab:r3dp-hparams}

\end{minipage}

\end{table*}


\subsection{R3DP Policy Training Details}
We adopt Diffusion Policy (DP) as the underlying visuomotor backbone and replace its original visual encoder with our 3D-aware perception encoder, which consists of the AFSC module, the pre-trained TFPNet, and the MVFF module. During policy training, both the VGGT encoder and the pretrained TFPNet remain frozen. Only the diffusion policy backbone is updated. Each training sample is a short trajectory segment consisting of 8 frames that are uniformly spaced by a stride of 8 control steps, resulting in an effective temporal window of 64 steps. For each frame, we use both head and front camera images together with proprioceptive states as the observation, and condition the diffusion denoiser on the fused 3D-aware features produced by R3DP. During policy training, the diffusion denoiser is conditioned on the same AFSC-generated features as used at inference time: VGGT features are computed on sparse key frames, while features for the remaining frames are provided by the frozen TFPNet.


We train a separate policy for each task using the same training configuration. This protocol is shared by both RoboTwin simulation and real-world experiments. The training hyperparameters are summarized in Table~\ref{tab:r3dp-hparams}.

\subsection{Details for Baselines and Ablation Variants}
To ensure a fair comparison, all Diffusion-Policy-based baselines and ablations are trained using the same demonstration sets, action horizon, and optimization hyperparameters as our full R3DP model (see Table~\ref{tab:r3dp-hparams}). For each method, we evaluate checkpoints at 300, 450, and 600 epochs under identical settings and report the best-performing checkpoint in the main tables. The architectural definitions of DP-single, DP-multi, DP+VGGT, DP+VGGT+MVFF, and R3DP with AFSC strictly follow the descriptions in the main paper, and no additional tuning beyond the shared configuration is applied.

Finally, to help readers better understand the structure of MVFF and its corresponding ablation, we provide simplified PyTorch-style pseudocode for both modules as follows. 

\begin{lstlisting}[language=Python,
caption={},
label={lst:mvff-code},
basicstyle=\footnotesize\ttfamily,
captionpos=b,
frame=none]
# Fusion; set geom=False for ablation w/o MVFF
class Fusion(nn.Module):
    def forward(self, view_tokens, Ks=None, Tcw=None, geom=True):
        x = flatten(view_tokens)
        q, k, v = proj(x)
        if geom:
            # MVFF: PRoPE + cam geo
            cam = flatten_cam(Ks, Tcw)
            a = prope_attn(to_heads(q,k,v), cam)
        else:
            # Abl.: standard dot-product attention
            a = sdpa(to_heads(q,k,v))
        y = from_heads(a)
        return unflatten(y)
\end{lstlisting}

%% file: main.bib
@String(CVPR  = {IEEE Conf. Comput. Vis. Pattern Recog.})

@String(ECCV  = {Eur. Conf. Comput. Vis.})

@String(ICLR  = {Int. Conf. Learn. Represent.})

@String(CVPR  = {CVPR})

@String(ECCV  = {ECCV})

@String(ICLR  = {ICLR})

@article{zhao2023learning,
  title={Learning fine-grained bimanual manipulation with low-cost hardware},
  author={Zhao, Tony Z and Kumar, Vikash and Levine, Sergey and Finn, Chelsea},
  journal={arXiv preprint arXiv:2304.13705},
  year={2023}
}

@inproceedings{black2025pi_,
  title={$\pi0.5$: a Vision-Language-Action Model with Open-World Generalization},
  author={Black, Kevin and Brown, Noah and Darpinian, James and Dhabalia, Karan and Driess, Danny and Esmail, Adnan and Equi, Michael Robert and Finn, Chelsea and Fusai, Niccolo and Galliker, Manuel Y and others},
  booktitle={9th Annual Conference on Robot Learning},
  year={2025}
}

@inproceedings{grotz2024peract2,
  title={Peract2: Benchmarking and learning for robotic bimanual manipulation tasks},
  author={Grotz, Markus and Shridhar, Mohit and Chao, Yu-Wei and Asfour, Tamim and Fox, Dieter},
  booktitle={CoRL 2024 Workshop on Whole-body Control and Bimanual Manipulation: Applications in Humanoids and Beyond},
  year={2024}
}

@inproceedings{shridhar2023perceiver,
  title={Perceiver-actor: A multi-task transformer for robotic manipulation},
  author={Shridhar, Mohit and Manuelli, Lucas and Fox, Dieter},
  booktitle={Conference on Robot Learning},
  pages={785--799},
  year={2023},
  organization={PMLR}
}

@article{lin2025sem,
  title={SEM: Enhancing Spatial Understanding for Robust Robot Manipulation},
  author={Lin, Xuewu and Lin, Tianwei and Huang, Lichao and Xie, Hongyu and Jin, Yiwei and Li, Keyu and Su, Zhizhong},
  journal={arXiv preprint arXiv:2505.16196},
  year={2025}
}

@article{li2025cameras,
  title={Cameras as relative positional encoding},
  author={Li, Ruilong and Yi, Brent and Liu, Junchen and Gao, Hang and Ma, Yi and Kanazawa, Angjoo},
  journal={arXiv preprint arXiv:2507.10496},
  year={2025}
}

@article{ke20243d,
  title={3d diffuser actor: Policy diffusion with 3d scene representations},
  author={Ke, Tsung-Wei and Gkanatsios, Nikolaos and Fragkiadaki, Katerina},
  journal={arXiv preprint arXiv:2402.10885},
  year={2024}
}

@article{ze2024generalizable,
  title={Generalizable humanoid manipulation with 3d diffusion policies},
  author={Ze, Yanjie and Chen, Zixuan and Wang, Wenhao and Chen, Tianyi and He, Xialin and Yuan, Ying and Peng, Xue Bin and Wu, Jiajun},
  journal={arXiv preprint arXiv:2410.10803},
  year={2024}
}

@article{ma2025cdp,
  title={CDP: Towards Robust Autoregressive Visuomotor Policy Learning via Causal Diffusion},
  author={Ma, Jiahua and Qin, Yiran and Li, Yixiong and Liao, Xuanqi and Guo, Yulan and Zhang, Ruimao},
  journal={arXiv preprint arXiv:2506.14769},
  year={2025}
}

@inproceedings{gong2025carp,
  title={Carp: Visuomotor policy learning via coarse-to-fine autoregressive prediction},
  author={Gong, Zhefei and Ding, Pengxiang and Lyu, Shangke and Huang, Siteng and Sun, Mingyang and Zhao, Wei and Fan, Zhaoxin and Wang, Donglin},
  booktitle={Proceedings of the IEEE/CVF International Conference on Computer Vision},
  pages={13460--13470},
  year={2025}
}

@article{miyato2023gta,
  title={Gta: A geometry-aware attention mechanism for multi-view transformers},
  author={Miyato, Takeru and Jaeger, Bernhard and Welling, Max and Geiger, Andreas},
  journal={arXiv preprint arXiv:2310.10375},
  year={2023}
}

@article{chi2025diffusion,
  title={Diffusion policy: Visuomotor policy learning via action diffusion},
  author={Chi, Cheng and Xu, Zhenjia and Feng, Siyuan and Cousineau, Eric and Du, Yilun and Burchfiel, Benjamin and Tedrake, Russ and Song, Shuran},
  journal={The International Journal of Robotics Research},
  volume={44},
  number={10-11},
  pages={1684--1704},
  year={2025}
}

@inproceedings{manuelli2019kpam,
  title={kpam: Keypoint affordances for category-level robotic manipulation},
  author={Manuelli, Lucas and Gao, Wei and Florence, Peter and Tedrake, Russ},
  booktitle={The International Symposium of Robotics Research},
  pages={132--157},
  year={2019}
}

@article{zeng2020transporter,
  title={Transporter Networks: Rearranging the Visual World for Robotic Manipulation},
  author={Zeng, Andy and Florence, Pete and Tompson, Jonathan and Welker, Stefan and Chien, Jonathan and Attarian, Maria and Armstrong, Travis and Krasin, Ivan and Duong, Dan and Wahid, Ayzaan and others},
  journal={arXiv preprint arXiv:2010.14406},
  year={2020}
}

@article{liu2024rdt,
  title={Rdt-1b: a diffusion foundation model for bimanual manipulation},
  author={Liu, Songming and Wu, Lingxuan and Li, Bangguo and Tan, Hengkai and Chen, Huayu and Wang, Zhengyi and Xu, Ke and Su, Hang and Zhu, Jun},
  journal={arXiv preprint arXiv:2410.07864},
  year={2024}
}

@article{lu2025h,
  title={H$^3$DP: Triply-Hierarchical Diffusion Policy for Visuomotor Learning},
  author={Lu, Yiyang and Tian, Yufeng and Yuan, Zhecheng and Wang, Xianbang and Hua, Pu and Xue, Zhengrong and Xu, Huazhe},
  journal={arXiv preprint arXiv:2505.07819},
  year={2025}
}

@inproceedings{Ze2024DP3,
	title={3D Diffusion Policy: Generalizable Visuomotor Policy Learning via Simple 3D Representations},
	author={Yanjie Ze and Gu Zhang and Kangning Zhang and Chenyuan Hu and Muhan Wang and Huazhe Xu},
	booktitle={Proceedings of Robotics: Science and Systems (RSS)},
	year={2024}
}

@inproceedings{o2024open,
  title={Open x-embodiment: Robotic learning datasets and rt-x models: Open x-embodiment collaboration 0},
  author={O’Neill, Abby and Rehman, Abdul and Maddukuri, Abhiram and Gupta, Abhishek and Padalkar, Abhishek and Lee, Abraham and Pooley, Acorn and Gupta, Agrim and Mandlekar, Ajay and Jain, Ajinkya and others},
  booktitle={2024 IEEE International Conference on Robotics and Automation (ICRA)},
  pages={6892--6903},
  year={2024},
  organization={IEEE}
}

@article{nair2022r3m,
  title={R3m: A universal visual representation for robot manipulation},
  author={Nair, Suraj and Rajeswaran, Aravind and Kumar, Vikash and Finn, Chelsea and Gupta, Abhinav},
  journal={arXiv preprint arXiv:2203.12601},
  year={2022}
}

@article{pearce2023imitating,
  title={Imitating human behaviour with diffusion models},
  author={Pearce, Tim and Rashid, Tabish and Kanervisto, Anssi and Bignell, Dave and Sun, Mingfei and Georgescu, Raluca and Macua, Sergio Valcarcel and Tan, Shan Zheng and Momennejad, Ida and Hofmann, Katja and others},
  journal={arXiv preprint arXiv:2301.10677},
  year={2023}
}

@article{qian2025gp3,
  title={GP3: A 3D Geometry-Aware Policy with Multi-View Images for Robotic Manipulation},
  author={Qian, Quanhao and Zhao, Guoyang and Zhang, Gongjie and Wang, Jiuniu and Xu, Ran and Gao, Junlong and Zhao, Deli},
  journal={arXiv preprint arXiv:2509.15733},
  year={2025}
}

@article{dinh2025improving,
  title={Improving Robotic Manipulation with Efficient Geometry-Aware Vision Encoder},
  author={Dinh Vuong, An and Nhat Vu, Minh and Reid, Ian},
  journal={arXiv e-prints},
  pages={arXiv--2509},
  year={2025}
}

@article{abouzeid2025geoaware,
  title={GeoAware-VLA: Implicit Geometry Aware Vision-Language-Action Model},
  author={Abouzeid, Ali and Mansour, Malak and Sun, Zezhou and Song, Dezhen},
  journal={arXiv preprint arXiv:2509.14117},
  year={2025}
}

@article{lin2025evo,
  title={Evo-0: Vision-Language-Action Model with Implicit Spatial Understanding},
  author={Lin, Tao and Li, Gen and Zhong, Yilei and Zou, Yanwen and Zhao, Bo},
  journal={arXiv preprint arXiv:2507.00416},
  year={2025}
}

@article{chen2025robotwin,
  title={RoboTwin 2.0: A Scalable Data Generator and Benchmark with Strong Domain Randomization for Robust Bimanual Robotic Manipulation},
  author={Chen, Tianxing and Chen, Zanxin and Chen, Baijun and Cai, Zijian and Liu, Yibin and Liang, Qiwei and Li, Zixuan and Lin, Xianliang and Ge, Yiheng and Gu, Zhenyu and others},
  journal={arXiv preprint arXiv:2506.18088},
  year={2025}
}

@InProceedings{fast3r,
    title={Fast3R: Towards 3D Reconstruction of 1000+ Images in One Forward Pass},
    author={Jianing Yang and Alexander Sax and Kevin J. Liang and Mikael Henaff and Hao Tang and Ang Cao and Joyce Chai and Franziska Meier and Matt Feiszli},
    booktitle={CVPR},
    year={2025},
}

@inproceedings{vggt,
  title={VGGT: Visual Geometry Grounded Transformer},
  author={Wang, Jianyuan and Chen, Minghao and Karaev, Nikita and Vedaldi, Andrea and Rupprecht, Christian and Novotny, David},
  booktitle={CVPR},
  year={2025}
}

@article{dens3r,
      title={Dens3R: A Foundation Model for 3D Geometry Prediction}, 
      author={Xianze Fang and Jingnan Gao and Zhe Wang and Zhuo Chen and Xingyu Ren and Jiangjing Lyu and Qiaomu Ren and Zhonglei Yang and Xiaokang Yang and Yichao Yan and Chengfei Lyu},
      journal={arXiv preprint arXiv:2507.16290},
      year={2025}
}

@article{MoRE,
  title={MoRE: 3D Visual Geometry Reconstruction Meets Mixture-of-Experts}, 
  author={Jingnan Gao and Zhe Wang and Xianze Fang and Xingyu Ren and Zhuo Chen and Shengqi Liu and Yuhao Cheng and Jiangjing Lyu and Xiaokang Yang and Yichao Yan},
  journal={arXiv preprint arXiv:2510.27234},
  year={2025}
}

@inproceedings{cut3r,
    Author = {Qianqian Wang and Yifei Zhang and Aleksander Holynski and Alexei A. Efros and Angjoo Kanazawa},
    Title = {Continuous 3D Perception Model with Persistent State},
    Year = {2025},
    booktitle={CVPR},
}

@inproceedings{flare,
title={FLARE: Feed-forward Geometry, Appearance and Camera Estimation from Uncalibrated Sparse Views}, 
      author={Shangzhan Zhang and Jianyuan Wang and Yinghao Xu and Nan Xue and Christian Rupprecht and Xiaowei Zhou and Yujun Shen and Gordon Wetzstein},
    Year = {2025},
      booktitle={CVPR},
}

@inproceedings{monst3r,
  author    = {Zhang, Junyi and Herrmann, Charles and Hur, Junhwa and Jampani, Varun and Darrell, Trevor and Cole, Forrester and Sun, Deqing and Yang, Ming-Hsuan},
  title     = {MonST3R: A Simple Approach for Estimating Geometry in the Presence of Motion},
    Year = {2025},
      booktitle={ICLR},
}

@inproceedings{feichtenhofer2019slowfast,
  title={Slowfast networks for video recognition},
  author={Feichtenhofer, Christoph and Fan, Haoqi and Malik, Jitendra and He, Kaiming},
  booktitle={Proceedings of the IEEE/CVF international conference on computer vision},
  pages={6202--6211},
  year={2019}
}

@inproceedings{mvdust3r,
  title={Mv-dust3r+: Single-stage scene reconstruction from sparse views in 2 seconds},
  author={Tang, Zhenggang and Fan, Yuchen and Wang, Dilin and Xu, Hongyu and Ranjan, Rakesh and Schwing, Alexander and Yan, Zhicheng},
  booktitle={CVPR},
  pages={5283--5293},
  year={2025}
}

@inproceedings{must3r,
  title={Must3r: Multi-view network for stereo 3d reconstruction},
  author={Cabon, Yohann and Stoffl, Lucas and Antsfeld, Leonid and Csurka, Gabriela and Chidlovskii, Boris and Revaud, Jerome and Leroy, Vincent},
  booktitle={CVPR},
  pages={1050--1060},
  year={2025}
}

@inproceedings{pow3r,
  title={Pow3r: Empowering unconstrained 3d reconstruction with camera and scene priors},
  author={Jang, Wonbong and Weinzaepfel, Philippe and Leroy, Vincent and Agapito, Lourdes and Revaud, Jerome},
  booktitle={CVPR},
  pages={1071--1081},
  year={2025}
}

@article{pi3,
title={$\pi^3$: Scalable Permutation-Equivariant Visual Geometry Learning}, 
      author={Yifan Wang and Jianjun Zhou and Haoyi Zhu and Wenzheng Chang and Yang Zhou and Zizun Li and Junyi Chen and Jiangmiao Pang and Chunhua Shen and Tong He},
      journal={arXiv preprint arXiv:2507.13347},
      year={2025}
}

@inproceedings{dust3r,
  author       = {Shuzhe Wang and Vincent Leroy and Yohann Cabon and Boris Chidlovskii and J{\'{e}}r{\^{o}}me Revaud},
  title        = {DUSt3R: Geometric 3D Vision Made Easy},
  booktitle    = {{CVPR}},
  pages        = {20697--20709},
  year         = {2024}
}

@inproceedings{mast3r,
  author       = {Vincent Leroy and
                  Yohann Cabon and
                  J{\'{e}}r{\^{o}}me Revaud},
  title        = {Grounding Image Matching in 3D with MASt3R},
  booktitle    = {{ECCV}},
  volume       = {15130},
  pages        = {71--91},
  year         = {2024}
}

@inproceedings{wang2024spann3r,
 title={3D Reconstruction with Spatial Memory},
  author={Wang, Hengyi and Agapito, Lourdes},
  booktitle    = {{3DV}},
  year         = {2025}
}

@article{shukor2025smolvla,
  title={Smolvla: A vision-language-action model for affordable and efficient robotics},
  author={Shukor, Mustafa and Aubakirova, Dana and Capuano, Francesco and Kooijmans, Pepijn and Palma, Steven and Zouitine, Adil and Aractingi, Michel and Pascal, Caroline and Russi, Martino and Marafioti, Andres and others},
  journal={arXiv preprint arXiv:2506.01844},
  year={2025}
}

@article{black2410pi0,
  title={$\pi$0: A vision-language-action flow model for general robot control},
  author={Black, Kevin and Brown, Noah and Driess, Danny and Esmail, Adnan and Equi, Michael and Finn, Chelsea and Fusai, Niccolo and Groom, Lachy and Hausman, Karol and Ichter, Brian and others},
  journal={arXiv preprint arXiv:2410.24164},
  year={2024}
}

@inproceedings{depthany,
  title={Depth anything: Unleashing the power of large-scale unlabeled data},
  author={Yang, Lihe and Kang, Bingyi and Huang, Zilong and Xu, Xiaogang and Feng, Jiashi and Zhao, Hengshuang},
  booktitle={Proceedings of the IEEE/CVF Conference on Computer Vision and Pattern Recognition},
  pages={10371--10381},
  year={2024}
}

@article{yang2024depth,
  title={Depth anything v2},
  author={Yang, Lihe and Kang, Bingyi and Huang, Zilong and Zhao, Zhen and Xu, Xiaogang and Feng, Jiashi and Zhao, Hengshuang},
  journal={Advances in Neural Information Processing Systems},
  volume={37},
  pages={21875--21911},
  year={2024}
}

@article{oquab2023dinov2,
  title={Dinov2: Learning robust visual features without supervision},
  author={Oquab, Maxime and Darcet, Timoth{\'e}e and Moutakanni, Th{\'e}o and Vo, Huy and Szafraniec, Marc and Khalidov, Vasil and Fernandez, Pierre and Haziza, Daniel and Massa, Francisco and El-Nouby, Alaaeldin and others},
  journal={arXiv preprint arXiv:2304.07193},
  year={2023}
}
